\documentclass{article} 
\usepackage{iclr2027_conference,times}

\usepackage{amsmath,amsfonts,bm}

\def\eqref#1{equation~\ref{#1}}

\def\1{\bm{1}}

\DeclareMathAlphabet{\mathsfit}{\encodingdefault}{\sfdefault}{m}{sl}
\SetMathAlphabet{\mathsfit}{bold}{\encodingdefault}{\sfdefault}{bx}{n}

\definecolor{linkblue}{RGB}{0,70,140}

\usepackage[
    colorlinks=true,
    linkcolor=linkblue,
    citecolor=linkblue,
    urlcolor=linkblue
]{hyperref}
\usepackage{url}
\usepackage{booktabs}
\usepackage{tabularx}
\usepackage{multirow}
\usepackage{graphicx}
\usepackage{amssymb}
\usepackage[table]{xcolor}
\usepackage{array}
\usepackage{enumitem}
\usepackage{tcolorbox}
\usepackage{fvextra}
\usepackage{fontawesome}
\usepackage{authblk}

\tcbuselibrary{breakable}

\definecolor{tsrbg}{RGB}{240,245,252}
\newcolumntype{C}{>{\centering\arraybackslash}X}
\newcolumntype{T}{>{\columncolor{tsrbg}\centering\arraybackslash}X}

\newcommand{\framework}{\text{IGMWorld}}

\newcommand{\bench}{\text{IGMBench}}

\title{World Editing: Intervening on Executable Worlds at Increasing Depth}

\author{
Max Ku$^{1,2}$ \quad
Nok-Kan Law$^{2}$ \quad
Yu-Chien Tang$^{2}$ \quad
Shih-Ying Yeh$^{2,3}$ \quad
Ping Nie$^{1}$ \newline
Andy Zheng$^{1}$ \quad
Tat Hei Lai$^{1}$ \quad
Fei-Yueh Chen$^{2}$ \quad
Nikko Yu$^{2}$ \quad
Wei-Chieh Sun$^{2}$ \newline
Suzy Huang$^{2}$ \quad
Chiao-Wei Hsu$^{2}$ \quad
Chih-Chuan Huang$^{2}$ \quad
Chak-Wing Mak$^{2}$ \newline
Ho Yin Sam Ng$^{2}$ \quad
Edisy Kin Wai Chan$^{2}$ \quad
Min-Hung Chen$^{2}$ \quad
Ho Kei Cheng$^{2,4}$\\
$^{1}$University of Waterloo \quad
$^{2}$G-G-G \quad
$^{3}$Comfy Org Research \quad
$^{4}$University of Illinois Urbana-Champaign \\
}

\iclrfinalcopy 
\begin{document}

\maketitle

\begin{abstract}
Interactive world models are increasingly capable of generating environments and acting within them, yet deliberately editing an existing executable world remains underexplored. We formulate \textbf{world editing} as intervening on an existing world while preserving properties that should remain unchanged, and introduce \textbf{intervention depth} as an axis describing how strongly an edit couples world entities, dynamics, and systems. We instantiate this capability through industry-grade game modding and introduce \framework{}, together with \bench{}, a benchmark of 110 tasks and over 1.1K executable state and behavioral criteria across Minecraft and Terraria. The tasks span property, entity, dynamics, and system interventions and are evaluated through deterministic executability, behavioral,
preservation, and visual checks. Frontier coding agents already exhibit substantial world-editing capability: the strongest configuration solves 78.2\% of tasks under a strict task-level criterion, while criterion-level performance reaches 94.8\%. Reliability generally decreases with intervention depth, and this pattern persists even among tasks with similar numbers of evaluation criteria. Most failed edits still build and load successfully, suggesting that the main difficulty is making the edited world behave as requested. Visual consistency remains a separate weakness, with all evaluated configurations below 50\% joint visual pass rate. These results show that world editing is a distinct capability from world generation and interaction, and that executable games provide a practical testbed for studying it.
\end{abstract}
\section{Introduction}


Contemporary research largely studies two relationships between AI systems and interactive worlds: generating worlds~\citep{bruce2024geniegenerativeinteractiveenvironments, guo2025mineworldrealtimeopensourceinteractive, valevski2025diffusionmodelsrealtimegame, solaris2026, hu2026multiplayerinteractiveworldmodels, marti2026opendreamer} and acting within worlds~\citep{fan2022minedojobuildingopenendedembodied, wang2023voyageropenendedembodiedagent, hafner2025trainingagentsinsidescalable, cheng2026agenticstsboundedmemorytestbedlonghorizon}. Many recent world models formulate generation primarily as predicting or synthesizing visual observations over time. Less studied is whether an AI system can deliberately edit an existing world: changing its entities, behaviors, or underlying systems while preserving unrelated properties. While prior work~\citep{mao2025yumeinteractiveworldgeneration} has used \textbf{world editing} for visual or geometric modifications of generated scenes, we use the term for \textbf{interventions} on an executable world itself, including its entities, dynamics, and interacting systems. We view world editing as a complementary capability to world generation and gameplay, and as a means of systematically constructing diverse, coherent world variants for training and evaluating future world models and agents.

This formulation raises two questions about the structure of world-editing capability. (1) \textbf{Are deeper interventions harder because they require coordinating more of the world, or simply because they contain more requirements?} (2) \textbf{When world edits fail, do they fail at basic executability, or after the game runs because the requested behavior is not correctly realized?} We instantiate these questions through industry-grade game modding\footnote{A \emph{mod} is a user-created modification of an existing game that changes or extends its assets, content, mechanics, or systems.}, where agents intervene on commercially released games through their native modding ecosystems. To study these questions, we introduce \framework{} and \bench{}. \framework{} provides an executable environment in which agents modify existing game repositories, build and run their edits, and are evaluated on the resulting world rather than on code structure alone. \bench{} contains 110 world-editing tasks across Minecraft~\citep{minecraft2009} and Terraria~\citep{relogic2011terraria}, spanning property, entity, dynamics, and system interventions, with over 1.1K executable state and behavioral evaluation criteria together with regression and visual validation. These games provide mature executable worlds whose entities, dynamics, assets, and systems can be programmatically modified and reproducibly evaluated~\citep{fabric2018,tmodloader}. We summarize our contributions as follows:
\begin{itemize}[noitemsep, topsep=2pt, leftmargin=2em]

\item We formulate \textbf{world editing} as intervention on an existing executable world and define \textbf{intervention depth} by the coupling among its entities, dynamics, and systems.

\item We operationalize world editing through \framework{} and \bench{}, comprising 110 tasks and over 1.1K executable evaluation criteria across Minecraft and Terraria, with deterministic state, behavioral, regression, and visual validation.

\item We use this testbed to characterize current frontier agents, finding that deeper edits are generally less reliable, most failed edits still build and load successfully but behave incorrectly, and visual integration remains a separate weakness.

\end{itemize}
\begin{table}[t]
\centering
\small
\setlength{\tabcolsep}{4pt}
\renewcommand{\arraystretch}{0.95}
\caption{
World-editing taxonomy and its realization in game modding.
}
\label{tab:mod_taxonomy}

\begin{tabularx}{\textwidth}{
    c
    >{\raggedright\arraybackslash}p{5cm}
    >{\raggedright\arraybackslash}p{4.2cm}
    >{\raggedright\arraybackslash}X
}
\toprule
\textbf{Level} &
\textbf{World Intervention} &
\textbf{Game Modding} &
\textbf{Examples} \\
\midrule

\textbf{L1} &
\textbf{Property Intervention}: Modify properties of existing components
without changing their identity or underlying behavior. &
\textbf{Parameter Editing}: Modify parameters, configurations, or
appearance of existing game components. &
Increase weapon damage;
replace a monster's visual asset. \\

\rule{0pt}{2ex}\textbf{L2} &
\textbf{Entity Intervention}: Introduce new entities or content while
largely preserving existing dynamics. &
\textbf{Content Editing}: Add new content that conforms to existing
mechanics. &
Add a crafting recipe or item;
add a monster or NPC. \\

\rule{0pt}{2ex}\textbf{L3} &
\textbf{Dynamics Intervention}: Modify rules governing entity behavior
and interaction. &
\textbf{Mechanic Editing}: Modify game logic governing behaviors and
interactions. &
Add an air-dash mechanic;
add a status-effect system. \\

\rule{0pt}{2ex}\textbf{L4} &
\textbf{System Intervention}: Modify multiple coupled components and
interactions forming a world-level system. &
\textbf{System Editing}: Modify multiple interacting game subsystems. &
Add a skill tree;
overhaul the economy system. \\

\bottomrule
\end{tabularx}
\end{table}

\begin{figure}[t]
\centering
\includegraphics[width=0.99\textwidth]{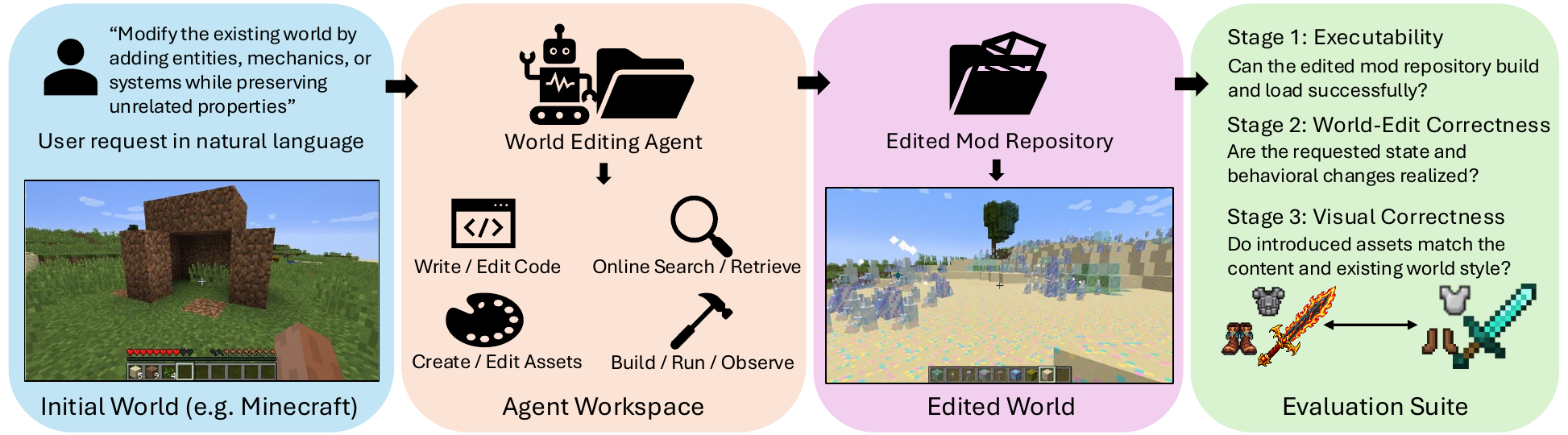}
\caption{
Overview of \framework{}. A world-editing agent modifies a scaffold mod
repository from a natural-language request and produces an edited world,
which is evaluated for executability, state and behavioral correctness,
and visual consistency with existing in-game assets.
}
\label{fig:pipeline}
\end{figure}

\section{World Editing as Intervention}

We formulate \textbf{world editing} as an \textbf{intervention} on an existing executable world. Let a world be abstractly represented as $W=(S,T,R,\ldots)$, where $S$ denotes its entities and state space, $T$ its transition dynamics, and $R$ the rules and systems that govern interactions. Given a natural-language editing instruction, a world-editing agent transforms the original world $W$ as 
\begin{equation}
    W \xrightarrow{\Delta_k} W' = f(W,\Delta_k),
    \qquad
    k \in
    \{\mathrm{property},\mathrm{entity},\mathrm{dynamics},\mathrm{system}\},
    \label{eq:world_edit}
\end{equation}
where $\Delta_k$ denotes an intervention of type $k$, and $W'$ is the resulting world that realizes the requested intervention while preserving unrelated properties of $W$. We characterize these interventions by their \textbf{depth of intervention}. Property interventions (L1) modify properties of existing world components without changing their identity or underlying behavior; Entity interventions (L2) introduce new entities or content while largely preserving existing world dynamics; Dynamics interventions (L3) introduce or modify rules governing how world entities behave and interact; and System interventions (L4) modify multiple coupled components and interactions that jointly constitute a world-level system. In game-modding terminology, these correspond respectively to parameter editing, content editing, mechanic editing, and system editing.

The levels characterize the semantic scope and coupling of the requested intervention rather than prescribing which implementation components must change. Intervention depth therefore measures how strongly an edit must coordinate multiple parts of the world, not how much code it requires: a system-level edit may be small in implementation yet still need several interacting components to remain consistent. Table~\ref{tab:mod_taxonomy} summarizes the four levels and representative examples. Games provide a useful testbed because their worlds are both complex and executable: entities, state transitions, rules, assets, and interacting systems are concretely instantiated through code and runtime behavior. Modding further exposes an intervention interface over an already existing world, rather than asking an agent to construct a world from scratch. Agents modify the implementation, but correctness is judged in the resulting executable world. Operationally, each task provides an initial repository, a requested intervention, an executable environment, and an independent validator. \framework{} provides the execution environment for these edits, while \bench{} provides the collection of task instances. Figure~\ref{fig:pipeline} illustrates this workflow.

\begin{table}[t]
\centering
\small
\caption{Distribution of tasks and evaluation criteria across games and
world-intervention levels. Each task specifies a natural-language world-edit
request and multiple state or behavioral criteria. A task succeeds only if
the implementation builds and passes all its criteria.}
\label{tab:benchmark_design_distribution}
\resizebox{\columnwidth}{!}{%
  \begin{tabular}{lrrrrrrrrrr}
  \toprule
  & \multicolumn{2}{c}{L1}
  & \multicolumn{2}{c}{L2}
  & \multicolumn{2}{c}{L3}
  & \multicolumn{2}{c}{L4}
  & \multicolumn{2}{c}{Total} \\
  \cmidrule(lr){2-3}
  \cmidrule(lr){4-5}
  \cmidrule(lr){6-7}
  \cmidrule(lr){8-9}
  \cmidrule(lr){10-11}
  Game
  & Tasks & Criteria
  & Tasks & Criteria
  & Tasks & Criteria
  & Tasks & Criteria
  & Tasks & Criteria \\
  \midrule
  Minecraft
  & 14 & 82
  & 15 & 144
  & 14 & 169
  & 14 & 226
  & 57 & 621 \\

  Terraria
  & 13 & 92
  & 14 & 162
  & 13 & 154
  & 13 & 132
  & 53 & 540 \\
  \midrule
  Total
  & 27 & 174
  & 29 & 306
  & 27 & 323
  & 27 & 358
  & 110 & \textbf{1161} \\
  \bottomrule
  \end{tabular}
}
\end{table}

\section{Operationalizing World Editing}
\label{sec:operation}

\textbf{Agent Execution Workflow.} \framework{} exposes each task as a reproducible $\text{edit} \rightarrow \text{build} \rightarrow \text{run} \rightarrow \text{observe}$ loop. The agent begins with a natural-language editing request and a prepared mod scaffold, and may inspect or modify source code and visual assets; invoke the native build toolchain; launch the game server and use logs to iteratively revise its implementation. The agent is given a fixed time budget of $T=3600$ seconds, after which the modified scaffold is submitted for evaluation. For each game, \framework{} provides a reproducible environment with the runtime, modding toolchain, required dependencies, and a fixed-seed world. Tasks within the same game share the same runtime and toolchain setup. For each task, the agent receives the request, its natural-language requirements, the project path, the build command, a generic build-and-load script, and constraints on immutable metadata. Executable verification checks are hidden from the agent. After execution, the modified scaffold is evaluated in a separate validation environment for build, load, and world-edit correctness.

\textbf{\bench{} Construction.}
We recruited 11 annotators with prior game-modding experience to author tasks across Minecraft and Terraria at all four intervention levels. Annotators were given the level definitions and representative examples, but were asked to write novel requests. Each task specifies the target game and modding ecosystem, an intervention level, a natural-language request, and evaluation criteria defining a correct implementation. Tasks were reviewed for clarity, feasibility, and level consistency, yielding 110 tasks. Each accepted task is compiled into an instruction view exposed to the agent and a verification view reserved for evaluation. The agent only sees the task description and a natural-language list of state and behavioral requirements. The corresponding executable checks are kept separate and used only during evaluation, so the agent never sees the task-specific grading logic. We instantiate \bench{} in Minecraft and Terraria using their native modding ecosystems, Fabric~\citep{fabric2018} and tModLoader~\citep{tmodloader}. The benchmark contains 110 tasks, with 57 in Minecraft and 53 in Terraria, distributed approximately evenly across the four intervention levels (Table~\ref{tab:benchmark_design_distribution}). In total, \bench{} contains around 1.1K evaluation criteria. The tasks span diverse gameplay domains, including combat, movement, entities, crafting, world generation, progression, and multi-system modifications; Appendix~\ref{app:diversity} reports the full distribution. The median number of criteria increases from 8 at L1 to 15 at L4. Because each task requires a native build, game launch, and multimodal validation, \bench{} is kept compact while still covering both games and all four intervention levels.

\subsection{Evaluation of Executable World Edits}
\label{sec:eval_pipeline}

\framework{} evaluates world edits along three complementary dimensions: \textbf{executability} asks whether the modified world can be built and loaded. \textbf{world-edit correctness} asks whether the requested intervention $\Delta_k$ is realized in the resulting world $W'$, including targeted checks that unrelated properties of $W$ are preserved. For edits involving visual assets, \textbf{visual correctness} asks whether the introduced content is semantically appropriate and visually
consistent with host world.

\textbf{Stage 1: Executability.} The submission must compile into a valid mod package and load successfully in the target game without crashing. Build and load checks are strict gates.

\textbf{Stage 2: World-Edit Correctness.}
For submissions that build and load successfully, \framework{} evaluates whether the requested intervention is realized through task-specific state and behavioral checks. State checks query properties of the running game, such as entity attributes, item statistics, recipes, and registrations.  Behavioral checks execute controlled in-game actions and verify the resulting state transitions. Where applicable, paired regression checks compare the modified and original worlds to test whether unrelated properties remain unchanged. These checks provide targeted rather than exhaustive evidence of preservation. \bench{} contains 225 regression checks across 30 tasks, concentrated at L1; coverage and pass rates are reported in Appendix~\ref{app:detail_results} (Table~\ref{tab:app_regression}). Validators assess observable world behavior rather than source-code structure, so different implementations may satisfy the same intervention. Additional details of the game-specific validation infrastructure are provided in Appendix~\ref{app:detail_setup}.

\textbf{Stage 3: Visual Consistency.}
For tasks that introduce or modify visual assets, \framework{} separately evaluates \textbf{semantic consistency}, whether the asset depicts the requested content, and \textbf{style consistency}, whether it matches the visual language of the host world. Both checks use TPIPS~\citep{wang2026tpips}, a text-conditioned extension of learned perceptual metrics such as LPIPS~\citep{zhang2018unreasonableeffectivenessdeepfeatures}, under different text factors. Let $x$ denote the evaluated asset and $\mathcal{I}_c=\{r_i\}$ the reference assets from category $c$. For a text factor $f$, let $\mathrm{TPIPS}(x,r_i;f)$ denote the similarity between $x$ and reference $r_i$, and let $\operatorname{kNN}(x,\mathcal{I}_c;f)$ denote the $k$ nearest category-matched references under the same factor. 
We define the text-conditioned consistency score $S_f(x;c)$ as
\begin{equation} 
S_f(x;c) = \frac{1}{k} \sum_{r_i \in \operatorname{kNN}(x,\mathcal{I}_c;\,f)} \mathrm{TPIPS}(x,r_i;f). 
\label{eq:tpips-knn} 
\end{equation} 
For semantic consistency, the text factor $f$ specifies the requested
semantic identity or object class; for style consistency, we use the factor
``art style''. For each category--factor pair, we calibrate a threshold
$\tau_{c,f}$ from the $\alpha$-quantile of leave-one-out scores over the
native reference set $\mathcal{I}_c$. An asset passes the corresponding
check when $S_f(x;c) \geq \tau_{c,f}$. Category conditioning avoids
comparisons across semantically incompatible assets, while $k$-NN
aggregation accommodates legitimate variation within a category.
Appendices~\ref{app:visual_metric_ablation} and~\ref{app:detail_setup}
provide calibration details, sensitivity analysis, human validation, and
comparison with CSD~\citep{somepalli2024measuringstylesimilaritydiffusion}.

\section{Characterizing World-Editing Capability}

\begin{table}[t]
\centering
\small
\caption{
Criterion Pass Rate (CPR) and World-Editing Success Rate (WSR) on \bench{},
broken down by world-intervention level.
CPR measures the fraction of individual state and behavioral criteria
satisfied, while WSR measures the fraction of world-editing tasks for
which all associated criteria are satisfied. Results are aggregated across Minecraft and Terraria. Visual correctness is evaluated separately in Table~\ref{tab:visual_results}.
}
\label{tab:results_by_level}

\begin{tabular}{lr>{\columncolor{tsrbg}}r
                r>{\columncolor{tsrbg}}r
                r>{\columncolor{tsrbg}}r
                r>{\columncolor{tsrbg}}r
                r>{\columncolor{tsrbg}}r}
\toprule
& \multicolumn{2}{c}{\textbf{L1}}
& \multicolumn{2}{c}{\textbf{L2}}
& \multicolumn{2}{c}{\textbf{L3}}
& \multicolumn{2}{c}{\textbf{L4}}
& \multicolumn{2}{c}{\textbf{Overall}} \\
\cmidrule(lr){2-3}
\cmidrule(lr){4-5}
\cmidrule(lr){6-7}
\cmidrule(lr){8-9}
\cmidrule(lr){10-11}

\textbf{Model}
& \textbf{CPR} & \textbf{WSR}
& \textbf{CPR} & \textbf{WSR}
& \textbf{CPR} & \textbf{WSR}
& \textbf{CPR} & \textbf{WSR}
& \textbf{CPR} & \textbf{WSR} \\
\midrule

DeepSeek-V4-Pro
& 92.1 & 74.1
& 80.8 & 24.1
& 47.7 & 18.5
& 47.0 & 11.1
& 63.3 & 31.8 \\

GLM-5.3
& 97.6 & 92.6
& 82.7 & 62.1
& 39.8 & 25.9
& 39.3 & 11.1
& 60.1 & 48.2 \\

Kimi-K3
& 98.5 & 88.9
& 78.3 & 65.5
& 61.0 & 51.9
& 73.3 & 33.3
& 74.9 & 60.0 \\

Gemini 3.5 Flash
& 87.7 & 77.8
& 86.4 & 55.2
& 83.5 & 37.0
& 68.8 & 59.3
& 81.0 & 57.3 \\

Claude Opus 4.8
& 99.6 & 92.6
& 92.8 & 72.4
& 94.7 & 66.7
& \textbf{89.4} & 55.6
& 93.6 & 71.8 \\

GPT-5.6 Luna
& 96.3 & 92.6
& 90.2 & 58.6
& 90.4 & 44.4
& 80.0 & \textbf{63.0}
& 88.5 & 64.5 \\

GPT-5.6 Sol
& \textbf{99.8} & \textbf{96.3}
& \textbf{93.9} & \textbf{86.2}
& \textbf{98.0} & \textbf{81.5}
& 88.8 & 48.1
& \textbf{94.8} & \textbf{78.2} \\

\bottomrule
\end{tabular}
\end{table}

\subsection{Experimental Setup}
\label{sec:experimental_setup}

\textbf{Agent and Environment setup.} We evaluate seven frontier agent configurations spanning proprietary and open-weight models. Proprietary models are evaluated using their standard or provider-recommended coding harnesses: GPT-5.6 Sol and GPT-5.6 Luna with Codex-CLI~\citep{OpenAI2026GPT56}, Gemini 3.5 Flash with Gemini-CLI~\citep{gemini35flash2026}, and Claude Opus 4.8 with Claude Code~\citep{frontier-claude-opus-4-8}. We therefore compare complete agent configurations rather than model backbones alone. For open-weight models, which do not share a canonical provider-specific coding harness, we evaluate DeepSeek-V4-Pro~\citep{deepseekai2026deepseekv4highlyefficientmilliontoken}, Kimi-K3~\citep{kimiteam2026kimik3openfrontier}, and GLM-5.3~\citep{glm5team2026glm5vibecodingagentic} using Hermes Agent~\citep{nousresearch_hermes_agent_2026}, a shared coding harness for all three models. All agents have access to Internet search and an optional image-generation tool: Codex-based configurations use gpt-image~\citep{openai2026chatgptimages}, while the others use nano-banana~\citep{googledeepmind2026nanobanana2} via MCP. Agents decide autonomously whether to use these tools. All configurations also receive the same generic game-specific modding skills and optional sprite post-processing pipeline (Appendix~\ref{app:ablations}), without task-specific or evaluator information. Unless otherwise noted, we use provider-recommended inference settings. The prompt is generated from the task specification and includes the request, natural-language requirements, project path, build command, a generic build-and-load script, and immutable-metadata constraints (Figure~\ref{fig:agent_prompt_example}). We provide no task-specific API guidance, worked examples, or executable validation logic. Each run has a wall-clock budget of $T=3600$ seconds, and each model--task pair is evaluated once. All runs use the pinned environments described in Section~\ref{sec:operation}: Minecraft~1.21.1 with Fabric and Java~21, and the pinned tModLoader release with its bundled .NET runtime for Terraria. Further configuration details are provided in Appendix~\ref{app:detail_setup} (Tables~\ref{tab:configurations} and~\ref{tab:docker}).

\textbf{Evaluation metrics.}
We report two metrics for functional world editing. Criterion Pass Rate (CPR) is the fraction of individual state and behavioral evaluation criteria satisfied across tasks. World-Editing Success Rate (WSR) is the fraction of tasks for which all associated state and behavioral criteria are satisfied. Thus, WSR is a strict task-level metric, whereas CPR captures partial correctness within a task. 
Visual consistency is evaluated separately for tasks involving introduced or modified assets. We report $\mathrm{Pass}_{\mathrm{style}}$, $\mathrm{Pass}_{\mathrm{sem}}$, and $\mathrm{Pass}_{\mathrm{both}}$, denoting the fractions of evaluated assets that pass the style, semantic, and both visual checks, respectively. 
Visual thresholds follow the calibration procedure in Section~\ref{sec:eval_pipeline}; calibration and sensitivity analysis are provided in Appendix~\ref{app:visual_eval}.

\begin{table}[t]
\centering
\small
\caption{
Visual quality of agent-produced assets.
$N$ counts registered texture assets; vanilla textures and missing files fail all checks.
$\mathrm{Pass}_{\mathrm{style}}$, $\mathrm{Pass}_{\mathrm{sem}}$, and
$\mathrm{Pass}_{\mathrm{both}}$ denote style, semantic, and joint pass rates, computed using the text-conditioned consistency score $S_f(x;c)$
in Equation~\ref{eq:tpips-knn} and the calibrated category-specific thresholds.
Vanilla (LOO) provides the in-game reference, with marginal style and semantic
pass rates of $1-\alpha=0.85$ by construction.
}
\label{tab:visual_results}

\setlength{\tabcolsep}{4pt}
\begin{tabular}{
l
ccc>{\columncolor{tsrbg}}c
ccc>{\columncolor{tsrbg}}c
}
\toprule
& \multicolumn{4}{c}{\textbf{Minecraft}}
& \multicolumn{4}{c}{\textbf{Terraria}} \\
\cmidrule(lr){2-5}
\cmidrule(lr){6-9}
\textbf{Model / Source}
& $N$
& $\mathrm{Pass}_{\mathrm{style}}$
& $\mathrm{Pass}_{\mathrm{sem}}$
& $\mathrm{Pass}_{\mathrm{both}}$
& $N$
& $\mathrm{Pass}_{\mathrm{style}}$
& $\mathrm{Pass}_{\mathrm{sem}}$
& $\mathrm{Pass}_{\mathrm{both}}$ \\
\midrule

Vanilla (LOO)
& 1989 & 0.85 & 0.85 & 0.80
& 15456 & 0.85 & 0.85 & 0.78 \\

\midrule

DeepSeek-V4-Pro
& 142 & 0.17 & 0.23 & 0.15
& 113 & 0.16 & 0.23 & 0.11 \\

GLM-5.3
& 226 & 0.49 & 0.53 & 0.40
& 73 & 0.32 & \textbf{0.37} & \textbf{0.29} \\

Kimi-K3
& 205 & \textbf{0.52} & \textbf{0.54} & \textbf{0.47}
& 109 & 0.29 & 0.35 & 0.24 \\

Gemini 3.5 Flash
& 110 & 0.32 & 0.36 & 0.22
& 104 & 0.32 & 0.32 & 0.22 \\

Claude Opus 4.8
& 127 & 0.34 & 0.42 & 0.23
& 110 & \textbf{0.35} & 0.34 & 0.25 \\

GPT-5.6 Luna
& 158 & 0.16 & 0.20 & 0.14
& 110 & 0.18 & 0.36 & 0.12 \\

GPT-5.6 Sol
& 155 & 0.23 & 0.24 & 0.19
& 99 & 0.22 & \textbf{0.37} & 0.18 \\


\bottomrule
\end{tabular}
\end{table}

\begin{figure}[t]
\centering
\begin{minipage}[t]{0.49\textwidth}
  \centering
  \includegraphics[width=\linewidth]{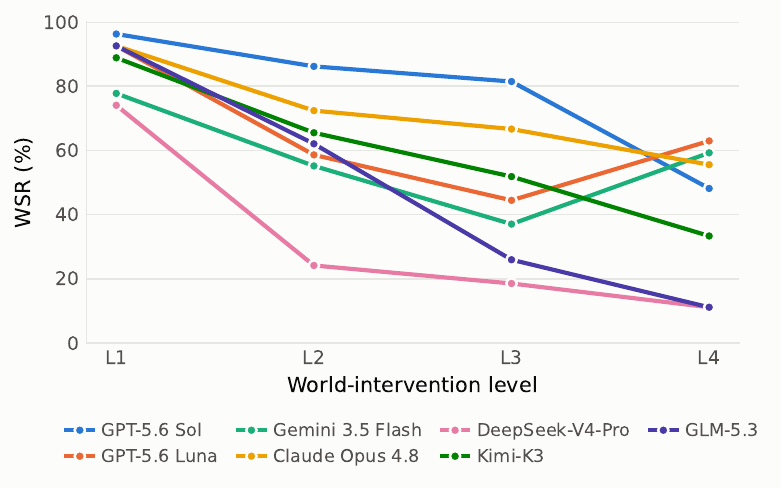}
  \par\smallskip
  (a) WSR by intervention level
\end{minipage}
\begin{minipage}[t]{0.49\textwidth}
  \centering
  \includegraphics[width=\linewidth]{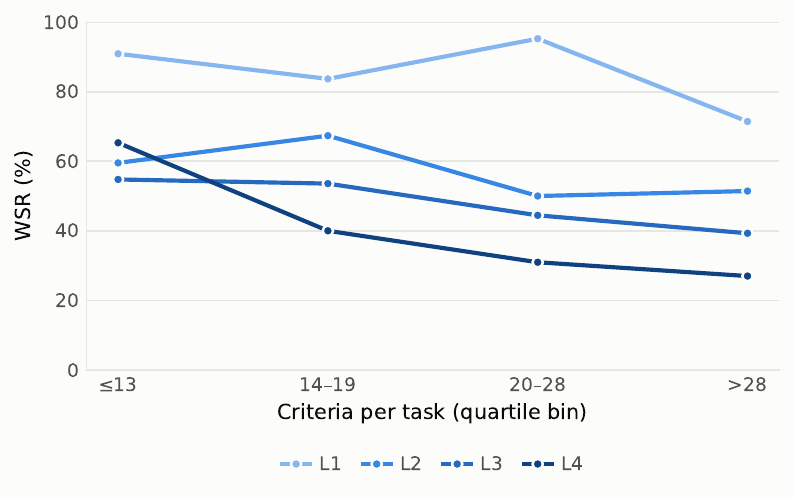}
  \par\smallskip
  (b) WSR controlled by criterion count
\end{minipage}
\caption{
Reliability across intervention depth. Performance differences persist within comparable criterion-count ranges, showing that intervention depth is not reducible to requirement count alone.
}
\label{fig:figureA}
\end{figure}

\subsection{Reliability Varies with Intervention Depth}
\label{sec:results_depth}

Table~\ref{tab:results_by_level} shows that current frontier agents already exhibit substantial functional world-editing capability. Under the strict World-Editing Success Rate (WSR), the evaluated configurations solve between 31.8\% and 78.2\% of \bench{} tasks, with GPT-5.6 Sol reaching 78.2\%. Criterion Pass Rate (CPR) is substantially higher, reaching 94.8\% for GPT-5.6 Sol, indicating that many unsuccessful tasks nevertheless satisfy substantial portions of the requested behavior. 

Reliability varies systematically with intervention depth. Every evaluated configuration achieves its highest WSR at L1, while performance is generally lower for deeper dynamics and system interventions, although the relative ordering of L3 and L4 varies across models and games. GPT-5.6 Sol, for example, decreases from 96.3\% WSR at L1 to 48.1\% at L4, while Claude Opus 4.8 decreases from 92.6\% to 55.6\%. At deeper levels, CPR remains substantially higher than WSR, indicating that agents often satisfy many requirements without completing the full edit. Because deeper tasks also contain more evaluation criteria, this trend could simply reflect the larger number of requirements that must be satisfied. We therefore stratify tasks by criterion count in Figure~\ref{fig:figureA}. Within comparable criterion-count ranges, L1 remains substantially more reliable, while L3 and L4 generally remain below shallower interventions. This suggests that intervention depth captures structure beyond criterion count alone.

\begin{figure}[t]
\centering
\begin{minipage}[t]{0.48\textwidth}
  \centering
  \includegraphics[width=\linewidth]{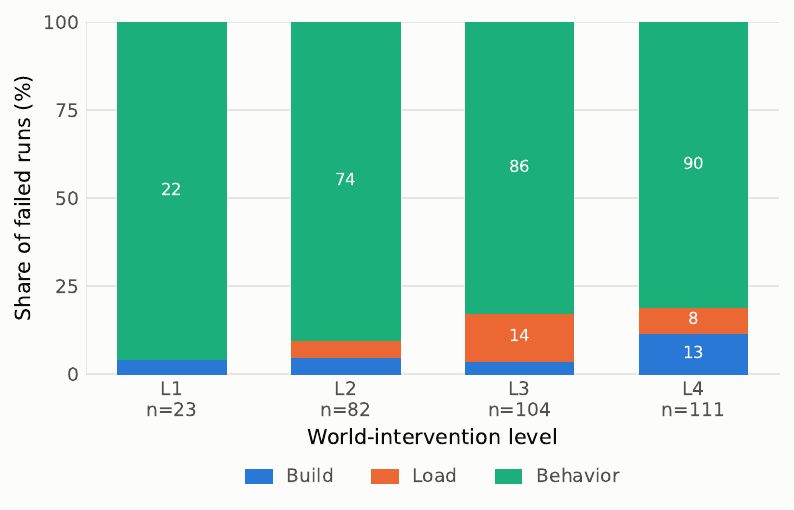}
  \par\smallskip
  (a) Failure stage by intervention level
\end{minipage}
\begin{minipage}[t]{0.48\textwidth}
  \centering
  \includegraphics[width=\linewidth]{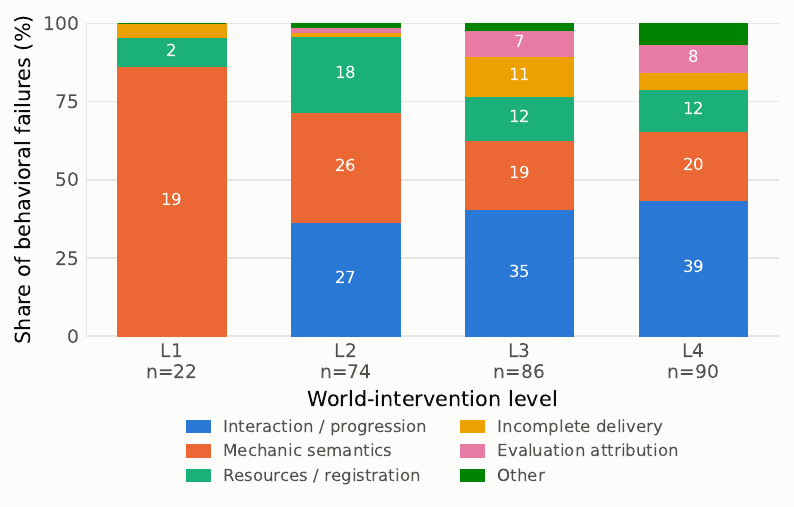}
  \par\smallskip
  (b) Behavioral failure domain by level
\end{minipage}
\caption{
Failure structure across intervention depth. Behavioral failures dominate across all levels. Among behavioral failures, interaction/progression failures increase from L2 to L4, while resource/registration failures remain relatively stable. Numbers inside bars denote failure counts.
}
\label{fig:figureB}
\end{figure}

\subsection{Most Failures Occur After Executability}
\label{sec:failure_depth}

Most unsuccessful world edits fail after reaching an executable state. Figure~\ref{fig:figureB}a shows that behavioral failures account for the large majority of unsuccessful runs at every intervention level, while build and load failures remain comparatively uncommon. Even at L4, where build failures become more frequent, behavioral correctness remains the dominant failure stage. This indicates that the principal bottleneck is not producing a modification that compiles and loads, but realizing the requested behavior in the executable world. The composition of behavioral failures also changes with intervention depth (Figure~\ref{fig:figureB}b). L1 failures are almost entirely mechanic semantics, such as a wrong value or formula. From L2 onward, interaction/progression failures, in which new content misbehaves in contact with existing systems, become the largest category and rise to 43\% at L4, while resource/registration failures remain at 13–24\% and do not grow with depth. The absence of interaction failures at L1 is expected by construction; the more informative pattern is the increase from L2 to L4 while resource/registration failures remain relatively stable. This is consistent with deeper edits requiring coordination across multiple interacting world components rather than a single localized change. Beyond correctness, preservation shows a different pattern. Targeted regression checks show few unintended changes on the tasks where such checks are available (Appendix~\ref{app:detail_results}, Table~\ref{tab:app_regression}), but coverage is concentrated at L1. Preservation under deeper interventions is therefore only partially tested and remains a limitation of the current
benchmark.

\begin{figure}[t]
\centering
\includegraphics[width=0.95\textwidth]{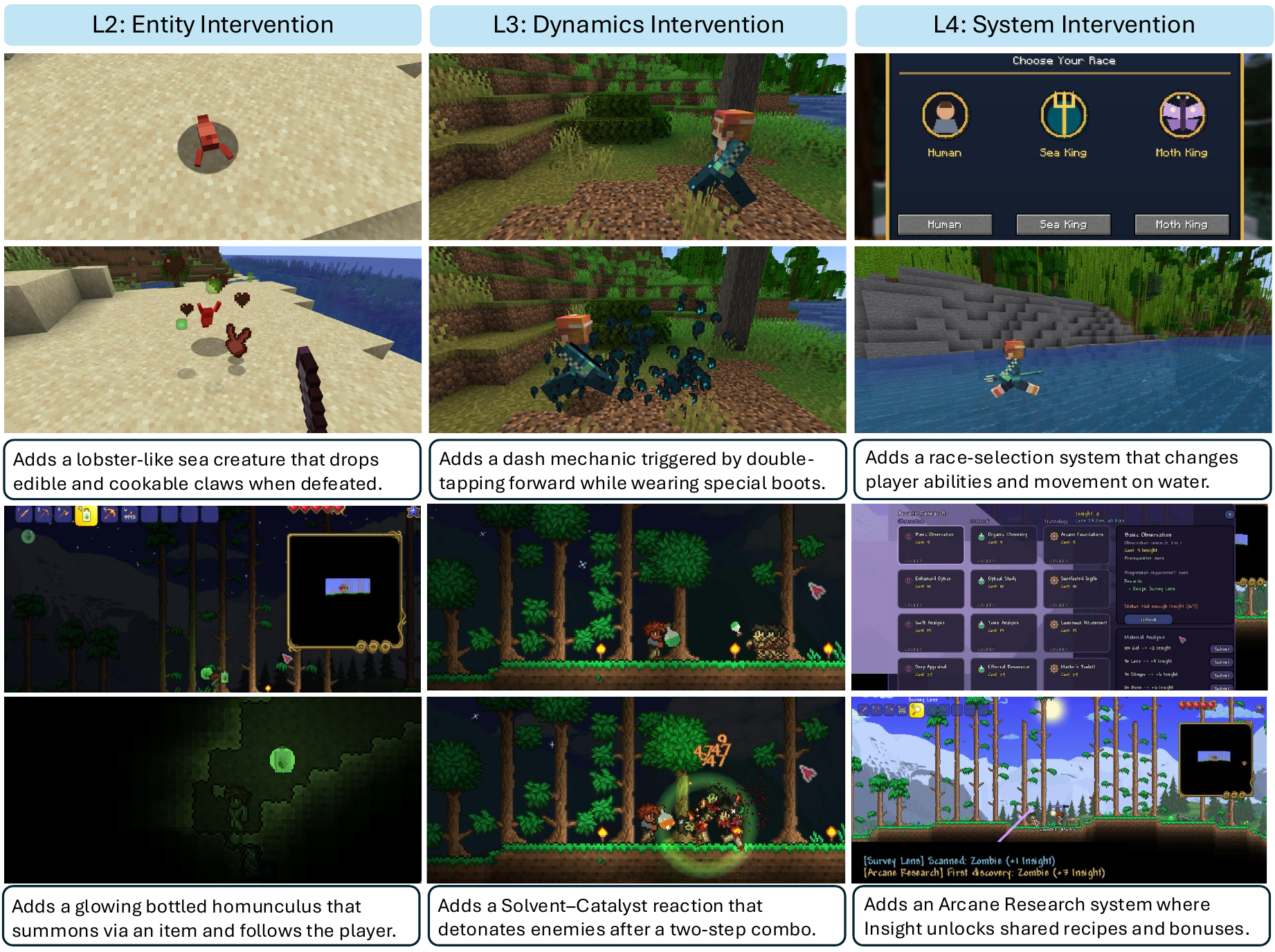}
\caption{
Representative successful world edits in Minecraft (top) and Terraria
(bottom) across L2--L4. From left to right, the examples introduce a new
entity, add new dynamics, and modify a coupled world-level system.
}
\label{fig:qual}
\end{figure}

\subsection{How Agents Edit Worlds}
\label{sec:agent_behavior}

Successful edits illustrate how intervention depth changes the structure of the required modification. As shown in Figure~\ref{fig:qual}, entity-level edits can often be realized by introducing localized content that conforms to existing world behavior, while dynamics-level edits require new state transitions or interaction rules. System-level edits additionally coordinate multiple components such as player state, interfaces, progression, and world behavior. These examples illustrate why intervention depth reflects coupling among world components rather than implementation size alone.

Agent trajectories reveal distinct strategies for grounding world edits. Table~\ref{tab:agent_behavior} shows that GPT-5.6 Sol, GPT-5.6 Luna, and Claude Opus 4.8 inspect existing game sources in 91--99\% of runs, while rebuilding after an initial implementation is common across configurations. Strategies nevertheless differ across harnesses: Gemini relies more heavily on online search, whereas GPT and Claude more often ground their implementations in local game sources, game data, and runtime logs. Visual tasks also expose different asset-construction strategies, including adapting existing assets, using image-generation tools, and programmatic drawing.

\subsection{Visual Consistency Is a Distinct Bottleneck}
\label{sec:visual_results}

Visual integration remains substantially weaker than functional world editing. Table~\ref{tab:visual_results} shows that native game assets achieve joint style-and-semantic pass rates of 0.80 in Minecraft and 0.78 in Terraria, whereas every evaluated agent configuration remains below 0.50. The strongest joint pass rates reach 0.47 in Minecraft and 0.29 in Terraria. Qualitative comparisons in Appendix Figures~\ref{fig:qual_assets_minecraft} and~\ref{fig:qual_assets_terraria} further illustrate how image-generated and programmatically drawn assets differ from native game art. Notably, visual performance does not follow the ranking observed for functional world editing. Configurations that are weaker on executable state and behavioral correctness can nevertheless perform competitively on visual assets, while the strongest functional agents do not dominate the visual evaluation. This weak correspondence suggests that world editing is multidimensional: successfully modifying the behavior of an executable world does not imply that newly introduced content integrates perceptually with that world.

\begin{figure}[t]
\centering
\includegraphics[width=0.99\textwidth]{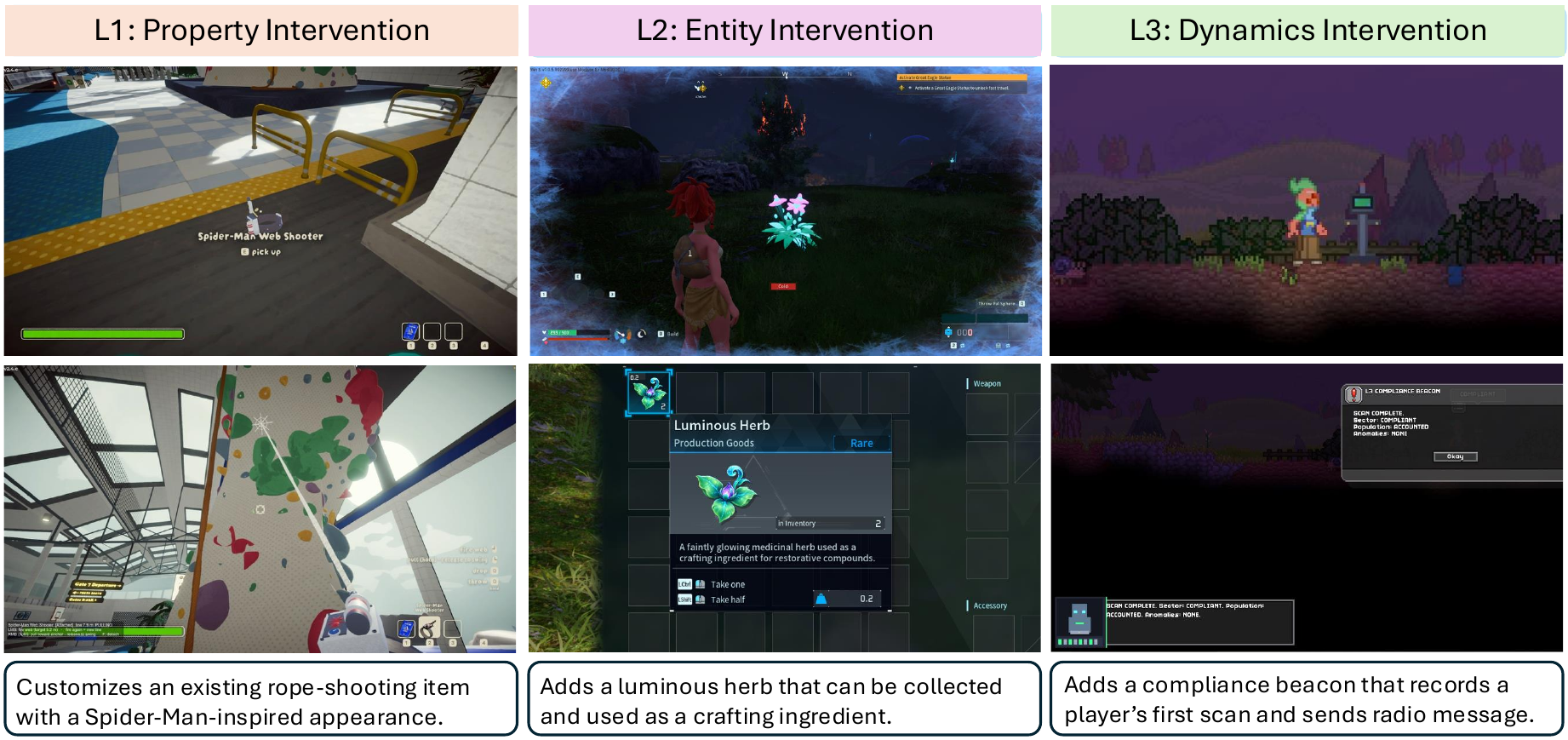}
\caption{
World editing beyond the games included in \bench{}. We show qualitative
examples of property, entity, and dynamics interventions in PEAK, Palworld,
and Starbound. These examples illustrate that the intervention taxonomy
applies to additional executable game worlds, but they are not 
part of the benchmark.
}
\label{fig:qual_extra}
\end{figure}

\section{Implications of World Editing}
\label{sec:implications}

World editing provides a mechanism for constructing controlled variants of an existing executable world while preserving much of its surrounding structure. This makes intervention depth a natural axis for organizing increasingly coupled changes, from localized property edits to modifications of dynamics and interacting systems. Such controlled variation could support future work on agent training, robustness, and generalization, although we do not evaluate these downstream uses here. World editing is also complementary to executable or code-based world models: rather than constructing or inferring a world representation, it asks whether an agent can deliberately intervene on an already instantiated world while maintaining consistency with what should remain unchanged. Although \bench{} quantitatively evaluates Minecraft and Terraria, the formulation is not specific to these games. Figure~\ref{fig:qual_extra} shows qualitative property, entity, and dynamics interventions in PEAK~\citep{peak_game_2025}, Palworld~\citep{palworld_2024}, and Starbound~\citep{starbound2016}. These examples suggest that world editing is better viewed as a general capability over executable worlds, with games serving as a concrete and verifiable testbed rather than the boundary of the problem.

\section{Related Work}

\textbf{World Models and Executable Worlds.} World models learn to represent, generate, or simulate an environment so that an agent can predict future states and interact over time~\citep{https://doi.org/10.5281/zenodo.1207631}. Recent research extends this idea to interactive 2D and 3D virtual worlds generated from text, images, or other multimodal inputs~\citep{alonso2024diffusionworldmodelingvisual, bruce2024geniegenerativeinteractiveenvironments, hunyuanworldteam2025hunyuanworld10generatingimmersive, hyworld2026hyworld20multimodalworld,
mao2025yumeinteractiveworldgeneration, mao2025yume15textcontrolledinteractiveworld, robbyantteam2026advancingopensourceworldmodels,gao2026infiniteworldsversatileinteractions}. A substantial portion of this literature focuses specifically on games, training models to predict environment dynamics from pixels, actions, or structured states~\citep{valevski2025diffusionmodelsrealtimegame,solaris2026, hu2026multiplayerinteractiveworldmodels, marti2026opendreamer, wang2026matrix, zhou2026agenticgamedevelopmentverifiable,li2026pixelsstatesrethinkinginteractive}. Most existing work focuses on generating, predicting, or simulating worlds or their behavior. World editing instead asks how an agent can deliberately intervene on an existing executable world while preserving what should remain unchanged. Recent work on code-based world models represents game or physical dynamics as executable programs, making states, transition rules, and governing mechanisms explicit and controllable~\citep{tang2024worldcodermodelbasedllmagent, lehrach2025codeworldmodelsgeneral, serapio2026distillinggamecodeworld, wang2026codeworldsagenticdiscovery, chen2026codeworldmodelcoding}. These works primarily construct or infer executable world representations, whereas world editing focuses on intervening on an already instantiated executable world. Our formulation further organizes such interventions by their depth, from localized property changes to modifications that couple multiple dynamics and systems.

\textbf{Agentic Game Development.} Game generation has long been studied as a content-creation problem, ranging from procedural level design~\citep{sarkar2020gamedesigncreativemachine, migdal2021levelgenerationstyleenhancement} to more recent generative model approaches that synthesize games or game assets~\citep{sudhakaran2023mariogptopenendedtext2levelgeneration, wang2023bytesized32corpuschallengetask, hu2024gamegenerationlargelanguage, li2025hunyuangameindustrialgradeintelligentgame, sani2026imagenworld}. In parallel, coding agents have progressed from isolated programming problems to repository-level software engineering, including bug fixing, feature implementation, testing, and large-scale codebase navigation~\citep{jimenez2024swebenchlanguagemodelsresolve, yang2024sweagentagentcomputerinterfacesenable, yang2024swebenchmultimodalaisystems}. Related benchmarks further evaluate agents that interact with software tools and graphical environments~\citep{xie2024osworldbenchmarkingmultimodalagents, huang2024blenderalchemyediting3dgraphics, yuan2026osworld20benchmarkingcomputer}. More recently, several works have applied agentic coding to game development, using code and engine APIs to construct playable games~\citep{ chen2025gamegptmultiagentcollaborativeframework, jiang2026opengameopenagenticcoding, zhang2026webgamebenchrequirementtoapplicationevaluationcoding, luo2026gamecraftbenchagentsbuildplayable, chi2026gamedevbenchevaluatingagenticcapabilities, yin2026autoueautomatedgeneration3d}. These efforts primarily focus on constructing games or completing game-development tasks. Closest to our setting, StarCharM~\citep{Zand_Miralvand_2025} studies generative AI for modding a commercial game (Stardew Valley), but as an HCI case study of a single content type without executable evaluation. In contrast, world editing treats modification of an existing executable world as the target capability. Success requires realizing the requested intervention while maintaining behavioral and visual consistency.

\section{Conclusion}

We introduced \textbf{world editing} as the capability to deliberately intervene on an existing executable world while preserving what should remain unchanged. We further proposed \textbf{intervention depth} as an axis describing how strongly a requested edit couples world entities, dynamics, and systems. Using executable games as a testbed, we find that current frontier coding agents exhibit substantial world-editing capability, but reliability varies strongly with intervention depth. This trend is not reducible to the number of evaluation criteria alone: deeper interventions remain less reliable within comparable criterion-count ranges. 
At the same time, most unsuccessful edits produce executable worlds yet fail during behavioral validation, 
indicating that the main bottleneck lies in realizing the requested world semantics rather than achieving basic build correctness. 
Visual integration remains a separate weakness, suggesting that functional and perceptual world editing are distinct dimensions of capability. Although \bench{} quantitatively evaluates Minecraft and Terraria, the underlying formulation is broader than game modding. 
Executable games provide a concrete setting in which interventions can be applied and verified, but the same notion of world editing extends to other executable worlds whose states, dynamics, and interacting systems can be modified. We hope \framework{} and \bench{} are a step toward treating world editing as a first-class capability alongside world generation and interaction: not only acting within or generating worlds, but deliberately transforming existing ones while maintaining their coherence.
\section*{Limitations}
\label{sec:limitations}

\textbf{Verification of open-ended world modifications.}
Even with task-specific state and behavioral criteria, no finite evaluator can exhaustively verify an arbitrary world modification. Some unintended side effects may remain outside the tested scenarios, particularly for deeper interventions involving multiple interacting systems. Broader verification would require stronger regression testing and more extensive runtime coverage, including long-horizon or multiplayer scenarios for interaction-dependent failures.

\textbf{Generalization across executable worlds.}
We instantiate \bench{} in Minecraft and Terraria using Fabric and tModLoader. Although these ecosystems are mature and support a broad range of interventions, other games expose different APIs, asset pipelines, runtime interfaces, and modding constraints. Figure~\ref{fig:qual_extra} illustrates that the same intervention taxonomy can be applied to additional games, including PEAK, Palworld, and Starbound, but these examples are not part of \bench{} and are not included in the quantitative evaluation. Quantitative evaluation across a broader range of executable worlds remains future work, as each environment requires ecosystem-specific execution and validation infrastructure.

\textbf{Beyond state, behavior, and visual consistency.}
Our evaluation primarily targets executable world state, behavior, and visual assets. Other modalities that contribute to a complete game experience, such as audio, animation, narrative, are not explicitly evaluated. Extending the benchmark to these dimensions is non-trivial: tasks involving such modalities are less naturally captured by deterministic state checks, and reliable automatic evaluation would require additional modality-specific infrastructure and validation protocols.

\textbf{Single-trial agent evaluation.}
Each model--task pair is evaluated once due to the cost of native game execution and multimodal validation. Consequently, our results characterize observed configuration-level performance rather than estimating per-task success probabilities across repeated stochastic runs. Repeated evaluation would provide stronger estimates of variance and reliability, but would substantially increase experimental cost.
\section*{Ethics statement}

The source code of \framework{}, the task specifications of \bench{}, and the accompanying evaluation scripts are original works of the authors. All benchmark tasks were authored from scratch. We did not copy, adapt, or redistribute files from existing publicly released mods, except for the official or community modding APIs and toolchains required to build and load mods. Minecraft and Terraria remain commercial works of their respective rights holders. Our release does not include game binaries, proprietary assets, or other copyrighted game files. Users must obtain the games and compatible toolchains independently and are responsible for complying with the applicable licenses and terms of use. The tasks in \bench{} are designed to evaluate benign world-editing capabilities and do not intentionally request harmful or deceptive modifications.

\section*{Acknowledgments}
We thank Yuntian Deng, Yu Ying Chiu, Zhi Rui Tam, Thomas Chong, Oscar Michel, and Jiarong Liang for helpful early-stage discussions and feedback.

\clearpage

\bibliography{iclr2027_conference}
\bibliographystyle{iclr2027_conference}

\clearpage

\appendix

\section{Appendix}

\subsection{More Representative Examples}
\label{app:representative_examples}

\begin{figure}[h]
\centering
\includegraphics[width=0.92\textwidth]{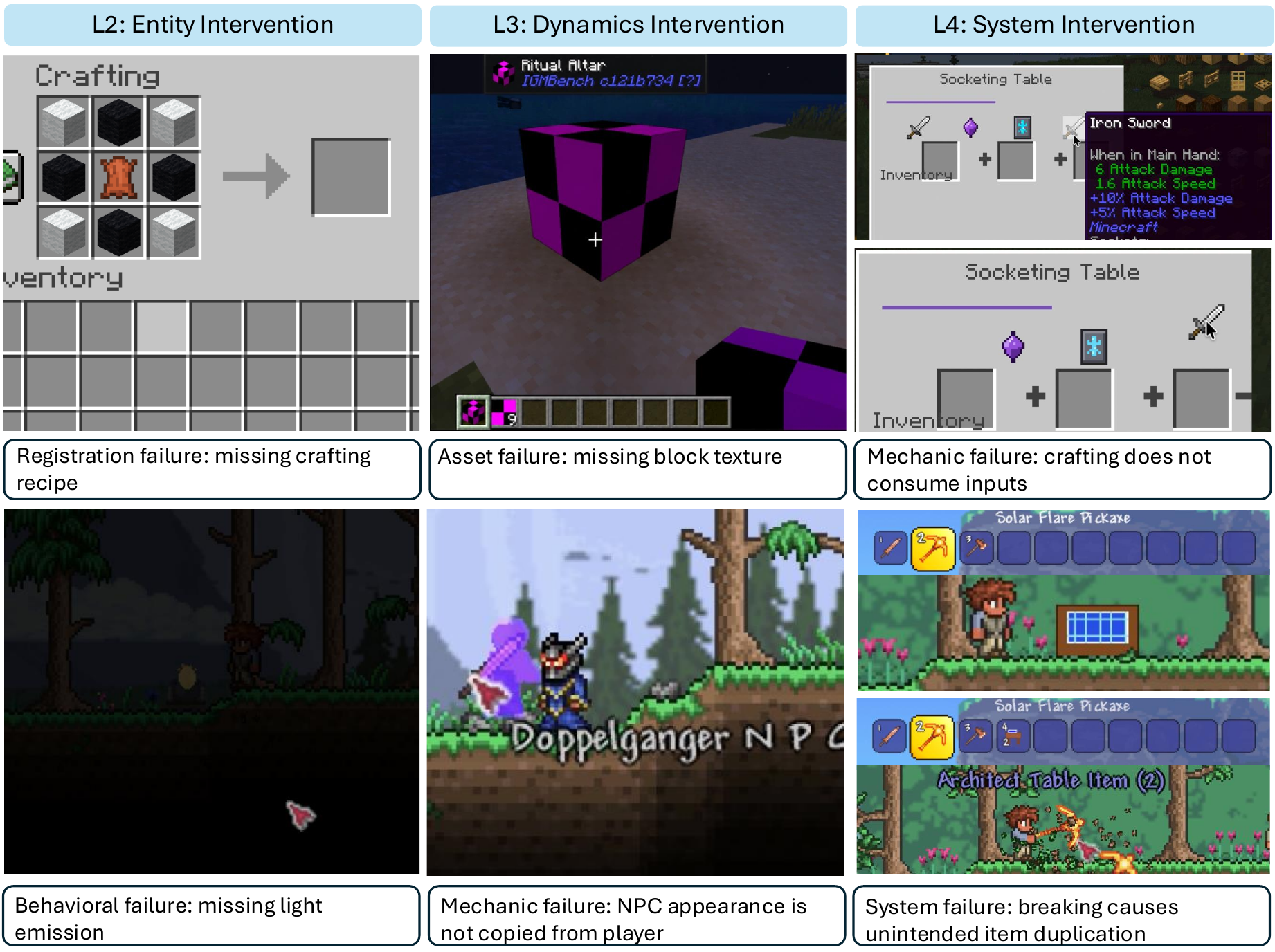}
\caption{
Representative failure cases across intervention levels in Minecraft
(top) and Terraria (bottom). Columns show entity (L2), dynamics (L3),
and system (L4) interventions. Failures include missing registration
or assets, incorrect mechanic execution, and unintended system-level
behavior.
}
\label{fig:qual_failure}
\end{figure}

\begin{figure}[h]
\centering
\includegraphics[width=0.98\textwidth]{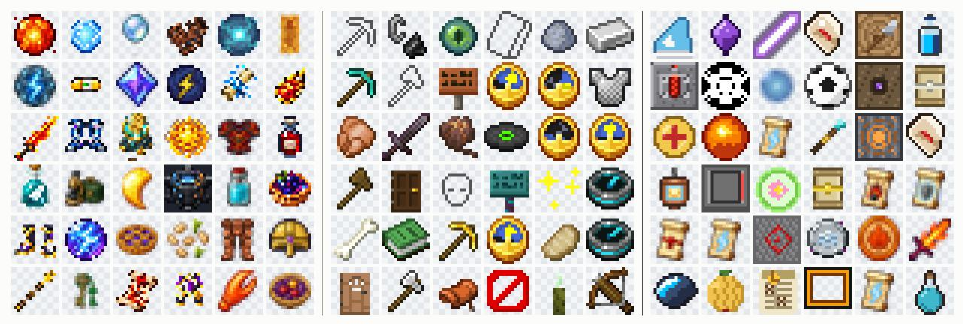}
\caption{
Minecraft item textures grouped by generation method:
\textbf{left}, image-generated; \textbf{middle}, native game assets;
\textbf{right}, programmatically drawn.
Image-generated assets tend toward smoother shading and higher contrast,
while programmatic assets are generally flatter and more geometric than
native Minecraft textures. Each panel shows 36 sampled item textures.
}
\label{fig:qual_assets_minecraft}
\end{figure}

\begin{figure}[h]
\centering
\includegraphics[width=0.98\textwidth]{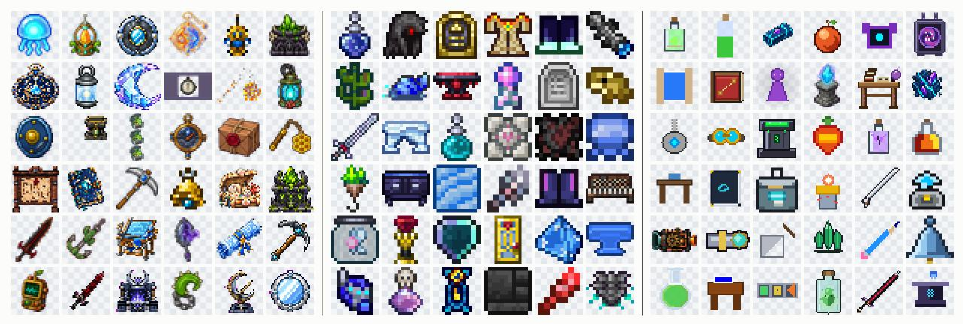}
\caption{
Terraria item textures grouped by generation method:
\textbf{left}, image-generated; \textbf{middle}, native game assets;
\textbf{right}, programmatically drawn.
Image-generated assets more closely reproduce Terraria's dense and outlined
visual style, while programmatic assets remain visibly simpler and more
geometric. Each panel shows sampled item textures from the same size range.
}
\label{fig:qual_assets_terraria}
\end{figure}

This section provides additional qualitative examples of failed world edits, together with visual comparisons of agent-generated assets. Figure~\ref{fig:qual_failure} shows representative failure cases across intervention levels. Figures~\ref{fig:qual_assets_minecraft} and~\ref{fig:qual_assets_terraria} compare image-generated and programmatically drawn assets against native game textures.

\section{Detailed Results}
\label{app:detail_results}

We provide detailed functional, visual, regression, and efficiency results in
Tables~\ref{tab:app_results_by_game_level}, \ref{tab:visual_results_full},
\ref{tab:app_regression}, and~\ref{tab:app_efficiency_game_level}, respectively.

\begin{table*}[h]
\centering
\scriptsize
\caption{
Performance on \bench{} across games and world-intervention levels.
Build Success Rate (BSR) measures the percentage of submissions that
build and load successfully. Criterion Pass Rate (CPR) measures the
percentage of individual state and behavioral criteria satisfied, with
criteria associated with non-building or non-loading submissions counted
as failed. World-Editing Success Rate (WSR) measures the percentage of
tasks that build and load successfully and satisfy all associated state
and behavioral criteria. WSR columns are shaded, and the highest reported
value in each column is shown in bold. All values are percentages.
}
\label{tab:app_results_by_game_level}
\setlength{\tabcolsep}{2pt}

\textbf{(a) Terraria}

\vspace{0.3em}
\begin{tabularx}{\textwidth}{
      >{\raggedright\arraybackslash}p{2.4cm}
      CCT CCT CCT CCT CCT
  }
  \toprule
  & \multicolumn{3}{c}{\textbf{L1}}
  & \multicolumn{3}{c}{\textbf{L2}}
  & \multicolumn{3}{c}{\textbf{L3}}
  & \multicolumn{3}{c}{\textbf{L4}}
  & \multicolumn{3}{c}{\textbf{Overall}} \\
  \cmidrule(lr){2-4}
  \cmidrule(lr){5-7}
  \cmidrule(lr){8-10}
  \cmidrule(lr){11-13}
  \cmidrule(lr){14-16}
  \textbf{Model}
  & BSR & CPR & WSR
  & BSR & CPR & WSR
  & BSR & CPR & WSR
  & BSR & CPR & WSR
  & BSR & CPR & WSR \\

  \midrule
  \multicolumn{16}{l}{\textit{Open-weight}} \\

  DeepSeek-V4-Pro
  & \textbf{100} & 94.9 & 76.9
  & \textbf{100} & 88.4 & 42.9
  & 61.5 & 33.0 & 23.1
  & 69.2 & 66.1 & 23.1
  & 83.0 & 68.9 & 41.5 \\

  GLM-5.3
  & \textbf{100} & \textbf{100} & \textbf{100}
  & 92.9 & 85.0 & 64.3
  & 84.6 & 13.5 & 23.1
  & 46.2 & 12.0 & 0.0
  & 81.1 & 53.0 & 47.2 \\

  Kimi-K3
  & \textbf{100} & 99.4 & 92.3
  & 85.7 & 71.5 & 64.3
  & 84.6 & 41.8 & 61.5
  & 84.6 & 69.3 & 53.8
  & 88.7 & 67.8 & 67.9 \\
  
  \midrule
  \multicolumn{16}{l}{\textit{Proprietary}} \\

Gemini 3.5 Flash
  & 92.3 & 86.1 & 76.9
  & \textbf{100} & 90.5 & 71.4
  & \textbf{100} & 88.8 & 46.2
  & \textbf{100} & \textbf{98.9} & 84.6
  & \textbf{98.1} & 90.6 & 69.8 \\

  Claude Opus 4.8
  & \textbf{100} & \textbf{100} & \textbf{100}
  & 92.9 & 93.0 & 78.6
  & \textbf{100} & \textbf{99.4} & \textbf{84.6}
  & \textbf{100} & 96.2 & 69.2
  & \textbf{98.1} & \textbf{97.0} & \textbf{83.0} \\

  GPT-5.6 Luna
  & \textbf{100} & 97.2 & 92.3
  & 92.9 & 92.2 & 64.3
  & 92.3 & 90.1 & 69.2
  & \textbf{100} & 97.9 & \textbf{92.3}
  & 96.2 & 93.7 & 79.2 \\

  GPT-5.6 Sol
  & \textbf{100} & \textbf{100} & \textbf{100}
  & \textbf{100} & \textbf{97.0} & \textbf{85.7}
  & \textbf{100} & 97.0 & 76.9
  & 92.3 & 90.0 & 61.5
  & \textbf{98.1} & 96.3 & 81.1 \\

  \bottomrule
  \end{tabularx}

\vspace{1em}

\textbf{(b) Minecraft}

\vspace{0.3em}

\begin{tabularx}{\textwidth}{
      >{\raggedright\arraybackslash}p{2.4cm}
      CCT CCT CCT CCT CCT
  }
  \toprule
  & \multicolumn{3}{c}{\textbf{L1}}
  & \multicolumn{3}{c}{\textbf{L2}}
  & \multicolumn{3}{c}{\textbf{L3}}
  & \multicolumn{3}{c}{\textbf{L4}}
  & \multicolumn{3}{c}{\textbf{Overall}} \\
  \cmidrule(lr){2-4}
  \cmidrule(lr){5-7}
  \cmidrule(lr){8-10}
  \cmidrule(lr){11-13}
  \cmidrule(lr){14-16}
  \textbf{Model}
  & BSR & CPR & WSR
  & BSR & CPR & WSR
  & BSR & CPR & WSR
  & BSR & CPR & WSR
  & BSR & CPR & WSR \\

  \midrule
  \multicolumn{16}{l}{\textit{Open-weight}} \\

  DeepSeek-V4-Pro
  & \textbf{100} & 85.6 & 71.4
  & 93.3 & 63.8 & 6.7
  & 78.6 & 65.2 & 14.3
  & 85.7 & 34.9 & 0.0
  & 89.5 & 56.0 & 22.8 \\

  GLM-5.3
  & \textbf{100} & 92.1 & 85.7
  & 93.3 & 77.6 & 60.0
  & 85.7 & 71.4 & 28.6
  & 92.9 & 56.8 & 21.4
  & 93.0 & 69.4 & 49.1 \\

  Kimi-K3
  & \textbf{100} & 96.4 & 85.7
  & \textbf{100} & \textbf{93.8} & 66.7
  & \textbf{100} & 83.7 & 42.9
  & \textbf{100} & 75.9 & 14.3
  & \textbf{100} & 84.1 & 52.6 \\
  
  \midrule
  \multicolumn{16}{l}{\textit{Proprietary}} \\

  Gemini 3.5 Flash
  & \textbf{100} & 91.4 & 78.6
  & 93.3 & 77.4 & 40.0
  & 85.7 & 77.1 & 28.6
  & 85.7 & 49.5 & 35.7
  & 91.2 & 68.5 & 45.6 \\

  Claude Opus 4.8
  & \textbf{100} & 98.6 & 85.7
  & \textbf{100} & 92.4 & 66.7
  & 92.9 & 89.2 & 50.0
  & \textbf{100} & 84.9 & \textbf{42.9}
  & 98.2 & 89.2 & 61.4 \\

  GPT-5.6 Luna
  & \textbf{100} & 94.2 & \textbf{92.9}
  & \textbf{100} & 85.7 & 53.3
  & \textbf{100} & 90.7 & 21.4
  & 85.7 & 68.2 & 35.7
  & 96.5 & 81.8 & 50.9 \\

  GPT-5.6 Sol
  & \textbf{100} & \textbf{99.3} & \textbf{92.9}
  & \textbf{100} & 86.7 & \textbf{86.7}
  & \textbf{100} & \textbf{99.2} & \textbf{85.7}
  & \textbf{100} & \textbf{88.1} & 35.7
  & \textbf{100} & \textbf{92.8} & \textbf{75.4} \\

  \bottomrule
  \end{tabularx}

\end{table*}

\begin{table}[t]
\centering
\small
\caption{
Detailed visual-quality results by game.
$N$ counts registered texture assets.
\emph{Custom} is the fraction using custom artwork rather than a vanilla
texture or missing file; non-custom assets fail all visual checks.
$S-\tau_c$ denotes the margin over the category-specific TPIPS threshold,
with positive values indicating a pass; \emph{Both Pass} requires both
style and semantic checks to pass.
}
\label{tab:visual_results_full}

\setlength{\tabcolsep}{3.5pt}
\begin{tabular}{lrcccccc}
\toprule
\textbf{Model / Source} &
$N$ &
\textbf{Custom} $\uparrow$ &
\textbf{Style} $S{-}\tau$ $\uparrow$ &
$\mathrm{Pass}_{\mathrm{style}}$ $\uparrow$ &
\textbf{Sem.} $S{-}\tau$ $\uparrow$ &
$\mathrm{Pass}_{\mathrm{sem}}$ $\uparrow$ &
$\mathrm{Pass}_{\mathrm{both}}$ $\uparrow$ \\
\midrule

\multicolumn{8}{l}{\textit{Minecraft} (item, block, entity)} \\
\midrule

Vanilla (LOO)
& 1989 & -- & +0.041 & 0.85 & +0.062 & 0.85 & 0.80 \\

\addlinespace

DeepSeek-V4-Pro
& 142 & 1.00 & -0.026 & 0.17 & -0.029 & 0.23 & 0.15 \\

GLM-5.3
& 226 & 0.94 & +0.001 & 0.49 & +0.003 & 0.53 & 0.40 \\

Kimi-K3
& 205 & 0.94 & -0.005 & 0.52 & -0.000 & 0.54 & 0.47 \\

Gemini 3.5 Flash
& 110 & 0.95 & -0.009 & 0.32 & -0.011 & 0.36 & 0.22 \\

Claude Opus 4.8
& 127 & 0.96 & -0.010 & 0.34 & -0.011 & 0.42 & 0.23 \\

GPT-5.6 Luna
& 158 & 0.97 & -0.027 & 0.16 & -0.030 & 0.20 & 0.14 \\


GPT-5.6 Sol
& 155 & 0.97 & -0.016 & 0.23 & -0.018 & 0.24 & 0.19 \\

\midrule

\multicolumn{8}{l}{
\textit{Terraria} (items and projectiles as sprites; tiles and NPCs as frames from vanilla sheets)
} \\
\midrule

Vanilla (LOO)
& 15456 & -- & +0.026 & 0.85 & +0.041 & 0.85 & 0.78 \\

\addlinespace

DeepSeek-V4-Pro
& 113 & 0.90 & -0.031 & 0.16 & -0.030 & 0.23 & 0.11 \\

GLM-5.3
& 73 & 0.74 & -0.012 & 0.32 & -0.003 & 0.37 & 0.29 \\

Kimi-K3
& 109 & 0.92 & -0.015 & 0.29 & -0.011 & 0.35 & 0.24 \\

Gemini 3.5 Flash
& 104 & 1.00 & -0.013 & 0.32 & -0.011 & 0.32 & 0.22 \\

Claude Opus 4.8
& 110 & 0.94 & -0.011 & 0.35 & -0.008 & 0.34 & 0.25 \\

GPT-5.6 Luna
& 110 & 0.85 & -0.019 & 0.18 & -0.004 & 0.36 & 0.12 \\


GPT-5.6 Sol
& 99 & 0.96 & -0.021 & 0.22 & -0.014 & 0.37 & 0.18 \\

\bottomrule
\end{tabular}
\end{table}

\begin{table}[h]
\centering
\small
\caption{
Regression preservation on \bench{}.
A regression check verifies that a property not targeted by the requested
edit remains unchanged, using a paired comparison between the modded world
and a mod-free world under the same environment.
\bench{} contains 225 regression checks across 30 tasks
(14 Minecraft and 16 Terraria), primarily at L1.
We report pass rates over checks reached by each configuration; runs that
fail to build or load do not execute regression checks.
}
\label{tab:app_regression}

\begin{tabular}{lr}
\toprule
\textbf{Model} & \textbf{Regression pass rate} \\
\midrule
\multicolumn{2}{l}{\textit{Open-weight}} \\
DeepSeek-V4-Pro    & 99.1\% \\
GLM-5.3            & 96.9\% \\
Kimi-K3            & 99.6\% \\
\midrule
\multicolumn{2}{l}{\textit{Proprietary}} \\
Gemini 3.5 Flash   & 98.2\% \\
Claude Opus 4.8    & 98.7\% \\
GPT-5.6 Luna       & 97.3\% \\
GPT-5.6 Sol        & \textbf{100.0\%} \\
\bottomrule
\end{tabular}
\end{table}

\begin{table}[t]
\centering
\scriptsize
\caption{Cost efficiency of evaluated agents across games and intervention levels. We report the median processed input tokens and agent runtime, where M denotes millions of tokens and m denotes minutes. The lowest value in each column is bolded.}
\label{tab:app_efficiency_game_level}
\setlength{\tabcolsep}{2.4pt}
\begin{tabular}{l*{16}{r}}
\toprule
& \multicolumn{8}{c}{Minecraft}
& \multicolumn{8}{c}{Terraria} \\
\cmidrule(lr){2-9}
\cmidrule(lr){10-17}
& \multicolumn{2}{c}{L1}
& \multicolumn{2}{c}{L2}
& \multicolumn{2}{c}{L3}
& \multicolumn{2}{c}{L4}
& \multicolumn{2}{c}{L1}
& \multicolumn{2}{c}{L2}
& \multicolumn{2}{c}{L3}
& \multicolumn{2}{c}{L4} \\
\cmidrule(lr){2-3}
\cmidrule(lr){4-5}
\cmidrule(lr){6-7}
\cmidrule(lr){8-9}
\cmidrule(lr){10-11}
\cmidrule(lr){12-13}
\cmidrule(lr){14-15}
\cmidrule(lr){16-17}
Model
& Token & Time & Token & Time & Token & Time & Token & Time
& Token & Time & Token & Time & Token & Time & Token & Time \\

\midrule
\multicolumn{17}{l}{\textit{Open-weight}} \\

DeepSeek-V4-Pro
& 1.3M & 8m & \textbf{2.5M} & 14m
& 3.8M & 19m & 4.4M & 19m
& 0.4M & 2m & 1.8M & 10m
& 2.7M & 13m & 5.4M & 20m \\

GLM-5.3
& 2.0M & 11m & 12.7M & 27m
& 11.3M & 27m & 12.9M & 31m
& 4.3M & 15m & 10.9M & 25m
& 11.3M & 27m & 10.6M & 27m \\

Kimi-K3
& 1.1M & 14m & 4.8M & 44m
& 6.6M & 47m & 6.8M & 38m
& 2.3M & 18m & 4.5M & 25m
& 9.5M & 32m & 9.6M & 35m \\

\midrule
\multicolumn{17}{l}{\textit{Proprietary}} \\

Gemini 3.5 Flash
& 1.7M & 16m & 2.8M & 26m
& 2.8M & 30m & \textbf{2.4M} & \textbf{13m}
& 0.7M & 4m & 1.8M & 18m
& \textbf{1.9M} & 14m & 5.3M & \textbf{13m} \\

Claude Opus 4.8
& \textbf{0.4M} & \textbf{3m} & \textbf{2.5M} & 11m
& \textbf{2.7M} & 14m & 4.3M & 18m
& \textbf{0.1M} & \textbf{1m} & \textbf{0.9M} & \textbf{6m}
& \textbf{1.9M} & \textbf{10m} & 4.6M & 19m \\

GPT-5.6 Sol
& 0.8M & \textbf{3m} & 5.2M & \textbf{10m}
& 5.1M & \textbf{13m} & 6.2M & 16m
& 0.2M & 2m & 2.6M & 9m
& 3.2M & 14m & \textbf{3.9M} & 19m \\

GPT-5.6 Luna
& 2.1M & 6m & 5.7M & 13m
& 8.1M & 15m & 6.9M & 16m
& 0.5M & 3m & 3.2M & 8m
& 5.5M & 14m & 7.1M & 19m \\

\bottomrule
\end{tabular}
\end{table}

\section{Failure Study and Agent Behaviors}

\begin{figure}[h]
\centering
\includegraphics[width=0.9\textwidth]{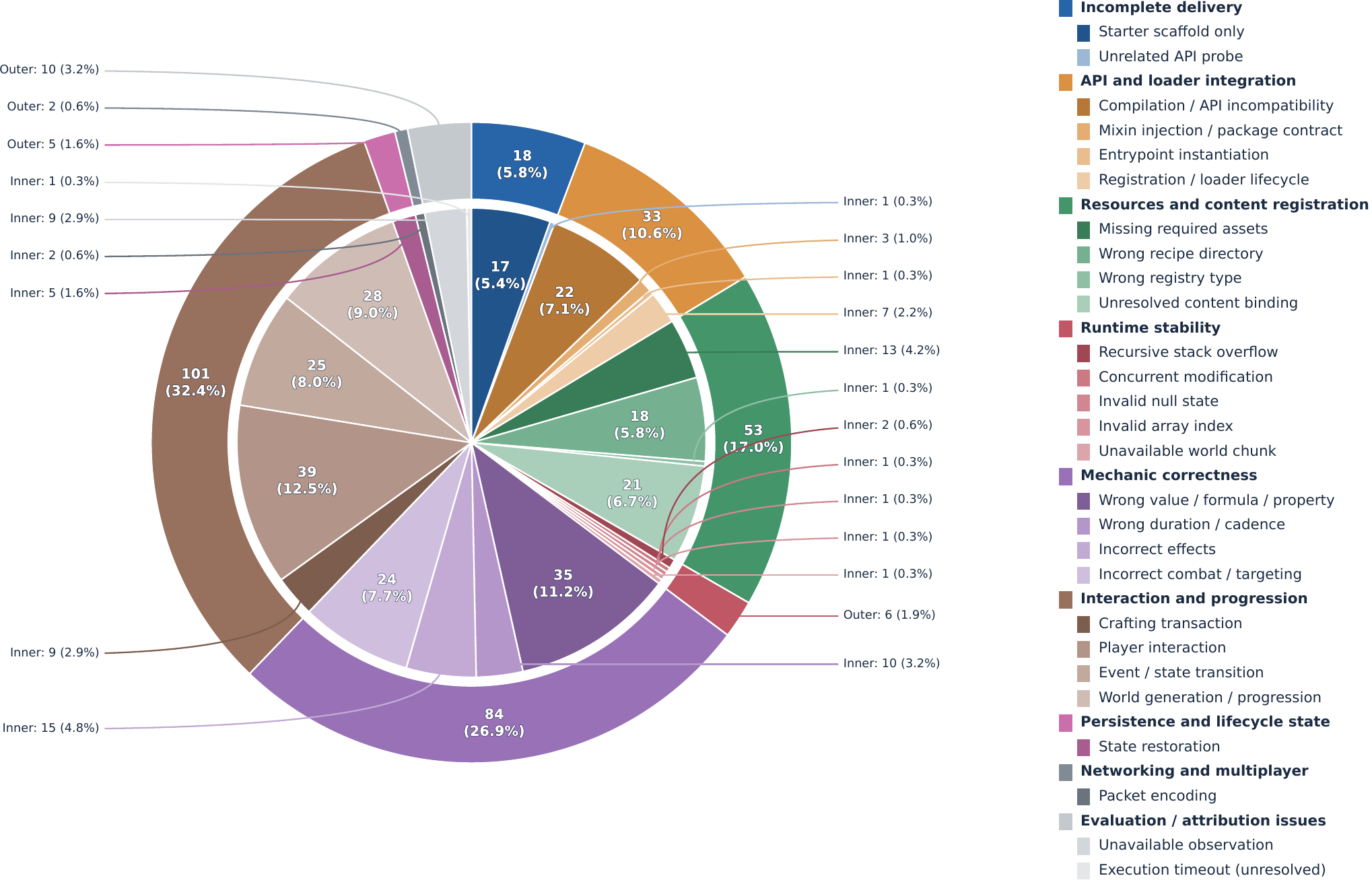}
\caption{Failure modes in agent-produced world edits. The outer ring
shows broad categories and the inner ring shows specific failure types;
labels report counts and percentages of categorized failures.}
\label{fig:failure_mode_pie}
\end{figure}

\textbf{Failure stages and domains.}
We further analyze unsuccessful and invalid runs to characterize where
world-editing attempts break down. Most failed submissions progress past
compilation and loading, with failures instead appearing during behavioral
validation (Table~\ref{tab:failure_stage}). Across the audited runs,
interaction/progression and mechanic semantics are the most common observed
failure domains (Figure~\ref{fig:failure_mode_pie};
Table~\ref{tab:failure_domains}). Figure~\ref{fig:failure_mode_pie} further
decomposes these broad domains into more specific failure types, showing that
unsuccessful world edits span both high-level behavioral errors and lower-level
integration failures. Because many cases remain unresolved with respect to
root cause, these categories describe where failures manifest rather than
necessarily identifying the underlying agent-side defect. Where the cause can
be established confidently, recurring errors are often integration-related,
including outdated resource conventions, registration-lifecycle mistakes,
missing assets, compilation failures, and incomplete delivery.

\textbf{Ecosystem-specific integration errors.}
Several recurring failures arise from incorrect assumptions about the target
game or modding ecosystem rather than from misunderstanding the requested
world edit. For example, multiple agents use outdated resource conventions
that produce otherwise plausible implementations but prevent the intended
content from being registered at runtime. Similar failures arise from
incorrect initialization order, loader/API assumptions, or resource layout.
These cases distinguish world-edit reasoning from ecosystem integration:
an agent may identify the correct entity, mechanic, or system-level change
but still fail to realize it in the running game because its implementation
does not conform to the target ecosystem.

\textbf{Development behavior and debugging.}
Trajectory analysis further shows that unsuccessful runs do not necessarily
correspond to less agent effort. For several strong configurations, failing
runs contain more tool use, probing, and log inspection than successful runs,
suggesting that difficult tasks often trigger longer debugging and repair
loops rather than immediate breakdowns. Other configurations exhibit the
opposite pattern, with failed runs showing less rebuilding and source
inspection. Across observable trajectories more generally, agents also adopt different
information-gathering strategies (Table~\ref{tab:agent_behavior}). GPT-5.6
Sol, GPT-5.6 Luna, and Claude Opus 4.8 predominantly inspect local game
sources, data files, and runtime logs, whereas Gemini 3.5 Flash relies much
more heavily on online search. Rebuilding after an initial implementation is
common across all four observable configurations, indicating that world
editing typically follows an iterative edit--build--inspect--repair workflow
rather than a single-shot generation process.

\begin{table}[h]
\centering
\small
\caption{
Observed failure stage by model across audited unsuccessful runs.
Counts indicate whether failures occur during build, load, or behavioral
validation.
}
\label{tab:failure_stage}
\begin{tabular}{lrrr}
\toprule
\textbf{Model} & \textbf{Build} & \textbf{Load} & \textbf{Behavior} \\
\midrule
DeepSeek-V4-Pro    & 2 & 13 & 60 \\
GLM-5.3            & 11 & 3 & 44 \\
Kimi-K3            & 5 & 1 & 39 \\
Gemini 3.5 Flash   & 4 & 2 & 42 \\
Claude Opus 4.8    & 0 & 2 & 27 \\
GPT-5.6 Luna       & 0 & 4 & 35 \\
GPT-5.6 Sol        & 0 & 1 & 25 \\
\bottomrule
\end{tabular}
\end{table}

\begin{table}[h]
\centering
\small
\caption{
Observed failure domains across audited unsuccessful runs.
These categories describe where failures are observed and do not always
identify the underlying root cause.
}
\label{tab:failure_domains}
\begin{tabular}{lr}
\toprule
\textbf{Observed domain} & \textbf{Count} \\
\midrule
Interaction / progression & 101 \\
Mechanic semantics         & 84 \\
Resources / registration   & 53 \\
API / loader               & 33 \\
Incomplete delivery        & 18 \\
Evaluation attribution     & 18 \\
Runtime crash              & 6 \\
Persistence                & 5 \\
Networking                 & 2 \\
\bottomrule
\end{tabular}
\end{table}

\begin{table}[t]
\centering
\small
\caption{
Agent development behavior across all \bench{} runs.
Values are the percentage of runs exhibiting each behavior.
Trajectory-level behaviors are reported only for harnesses whose
execution traces are observable.
}
\label{tab:agent_behavior}
\begin{tabular}{lrrrr}
\toprule
& \textbf{GPT-5.6 Sol}
& \textbf{GPT-5.6 Luna}
& \textbf{Gemini 3.5 Flash}
& \textbf{Claude Opus 4.8} \\
\midrule

Read game sources
& 97\% & 99\% & 52\% & 91\% \\

Read vanilla data
& 56\% & 52\% & 24\% & 37\% \\

Searched the Web
& 9\% & 9\% & 72\% & 3\% \\

Used provided skills
& 100\% & 96\% & 93\% & 58\% \\

Rebuilt after initial build
& 85\% & 90\% & 83\% & 76\% \\

Ran game server
& 15\% & 19\% & 0\% & 6\% \\

Ran own probe / test
& 30\% & 46\% & 13\% & 16\% \\

Read logs
& 66\% & 80\% & 27\% & 49\% \\

Kept a todo list
& 25\% & 65\% & 0\% & 0\% \\

\bottomrule
\end{tabular}
\end{table}

\clearpage
\section{Ablations}
\label{app:ablations}

\subsection{Use of Agent Skills}

We provide all evaluated agents with the same set of host-provided,
game-specific skills as part of the \framework{} execution environment.
These skills contain generic modding knowledge rather than task-specific
solutions or evaluator information. In particular,
\emph{\framework{}-minecraft-modding} summarizes common Fabric
registration patterns, client/server-side conventions, and asset and model
paths, while \emph{\framework{}-terraria-modding} provides corresponding
tModLoader class and registration patterns. For visual asset construction,
agents also have access to \emph{agent-sprite-forge}~\citep{agentSpriteForge},
an external pipeline for generating sprite sheets on a keyable background,
chroma-keying them locally, extracting frames, and exporting assets with
transparency. We also observe that, even without
\emph{agent-sprite-forge}, agents are often able to post-process their own
generated outputs and produce assets with transparent backgrounds, suggesting
that the pipeline provides convenience rather than an essential capability. The skills are made available uniformly, while agents decide autonomously
whether and how to use them. This is reflected in the observed trajectories:
provided skills are used in 58--100\% of runs across the proprietary
configurations with observable execution traces
(Table~\ref{tab:agent_behavior}). We include these skills in the default
configuration to provide the same generic ecosystem guidance across runs and
reduce incidental friction from modding-specific conventions, while
withholding task-specific implementation hints and evaluator information.

In this ablation, the no-skill condition removes both game-specific modding
skills and \emph{Agent Sprite Forge}, while leaving all other tools and
inference settings unchanged. To measure their effect, we rerun GPT-5.6 Luna without the host-provided
skills while keeping the remaining configuration fixed.
Table~\ref{tab:ablation_skills} shows that providing skills increases overall
WSR from 57.3\% to 64.5\%, with the largest gain at L4
(33.3\% to 63.0\%). The effect is not uniform across intervention levels:
L1 WSR is unchanged, L2 improves with skills, while L3 is higher without
skills. Overall CPR is nearly unchanged, at 88.5\% with skills and 89.2\%
without skills. This suggests that the skills primarily affect the
reliability of completing all requirements of a task rather than
criterion-level correctness in isolation. The skills do not provide a corresponding improvement in visual quality.
As shown in Table~\ref{tab:ablation_skills_visual}, the no-skill condition
achieves slightly higher style, semantic, and joint visual pass rates in both
Minecraft and Terraria. This indicates that the game-modding skills mainly
support functional implementation and ecosystem integration, while visual
asset quality remains governed by a different set of agent choices and tools.
We therefore use the skill-equipped configuration as the default setting in
the main experiments and report the no-skill configuration as a controlled
ablation.

\begin{table}[h]
\centering
\small
\caption{
Effect of host-provided skills on GPT-5.6 Luna.
We report Criterion Pass Rate (CPR) and World-Editing Success Rate (WSR)
across intervention levels, pooled over both Minecraft and Terraria. All values are percentages.
}
\label{tab:ablation_skills}

\begin{tabular}{lcccccccccc}
\toprule
& \multicolumn{2}{c}{\textbf{L1}}
& \multicolumn{2}{c}{\textbf{L2}}
& \multicolumn{2}{c}{\textbf{L3}}
& \multicolumn{2}{c}{\textbf{L4}}
& \multicolumn{2}{c}{\textbf{Overall}} \\
\cmidrule(lr){2-3}
\cmidrule(lr){4-5}
\cmidrule(lr){6-7}
\cmidrule(lr){8-9}
\cmidrule(lr){10-11}

\textbf{Configuration}
& CPR & WSR
& CPR & WSR
& CPR & WSR
& CPR & WSR
& CPR & WSR \\
\midrule

Luna (w/ skills)
& 96.3 & 92.6
& 90.2 & 58.6
& 90.4 & 44.4
& 80.0 & 63.0
& 88.5 & 64.5 \\

Luna (w/o skills)
& 97.8 & 92.6
& 89.8 & 44.8
& 93.3 & 59.3
& 78.5 & 33.3
& 89.2 & 57.3 \\

\bottomrule
\end{tabular}
\end{table}

\begin{table}[h]
\centering
\small
\caption{
Effect of host-provided skills on visual asset quality for GPT-5.6 Luna.
We report art-style, semantic, and joint visual pass rates for Minecraft
and Terraria.
}
\label{tab:ablation_skills_visual}
\setlength{\tabcolsep}{5pt}

\begin{tabular}{llccc}
\toprule
\textbf{Game} &
\textbf{Configuration} &
$\mathrm{Pass}_{\mathrm{style}}$ &
$\mathrm{Pass}_{\mathrm{sem}}$ &
$\mathrm{Pass}_{\mathrm{both}}$ \\
\midrule

Minecraft
& Luna (w/ skills)
& 0.16 & 0.20 & 0.14 \\

& Luna (w/o skills)
& 0.22 & 0.36 & 0.18 \\

\midrule

Terraria
& Luna (w/ skills)
& 0.18 & 0.36 & 0.12 \\

& Luna (w/o skills)
& 0.21 & 0.38 & 0.19 \\

\bottomrule
\end{tabular}
\end{table}

\subsection{Alternative Visual Asset Metrics}
\label{app:visual_metric_ablation}

\textbf{CSD as Style Metric.}
We additionally evaluate CSD~\citep{somepalli2024measuringstylesimilaritydiffusion}
as an alternative representation for measuring visual style consistency. Given a generated asset
$x$ and category-matched reference set $\mathcal{I}_c$, we embed each
image using the pretrained CSD encoder $f_{\mathrm{CSD}}$, a CLIP~\citep{radford2021learningtransferablevisualmodels}-based style embedding, and compute
the mean cosine similarity to the $k$ nearest references:
\begin{equation}
    S_{\mathrm{CSD}}(x;c)
    =
    \frac{1}{k}
    \sum_{r_i \in \operatorname{kNN}(\mathbf{z}_x,\mathcal{I}_c)}
    \cos(\mathbf{z}_x,\mathbf{z}_i),
\end{equation}
where
$\mathbf{z}_x=f_{\mathrm{CSD}}(x)/
\lVert f_{\mathrm{CSD}}(x)\rVert_2$
and $\mathbf{z}_i$ is defined analogously.

We calibrate category-specific thresholds using the same leave-one-out
protocol as the main TPIPS evaluator. To compare the underlying
representations independently of threshold calibration, we additionally
measure their threshold-free agreement with human style judgments.
Nine raters with prior gaming experience evaluated 40 blind-sampled
agent-produced textures and judged whether each asset matched the game's
art style. As shown in Table~\ref{tab:style_metric_validation}, TPIPS
conditioned on the factor ``art style'' achieves an AUC of 0.66 against
the majority human judgment, compared with 0.54 for CSD. We therefore use
TPIPS for style consistency in the main evaluation pipeline and retain CSD
as an alternative-metric comparison. The detail of human visual study is discussed in Appendix~\ref{app:visual_eval}.

\begin{table}[h]
\centering
\small
\caption{
Agreement with majority human judgments of whether agent-produced textures
match the game's art style, measured by threshold-free AUC.
Nine raters with prior gaming experience evaluated 40 blind-sampled textures;
unsure responses were excluded before majority voting.
}
\label{tab:style_metric_validation}

\begin{tabular}{lcc}
\toprule
& \textbf{CSD} & \textbf{TPIPS art style} \\
\midrule
\textbf{Majority vote}
& 0.54
& \textbf{0.66} \\
\bottomrule
\end{tabular}
\end{table}

\clearpage
\section{Detail Setup}
\label{app:detail_setup}

\subsection{Agent and Environment Configuration}
\label{app:agent_setup}

\begin{table}[h]
\centering
\small
\caption{
Agent configurations evaluated on \bench{}.
Each configuration specifies a backbone model, coding harness, inference
effort, and image-generation backend.  All models support approximately 1M-token context windows.
}
\label{tab:configurations}
\begin{tabular}{@{}llccc@{}}
\toprule
\textbf{Model} & \textbf{Harness} & \textbf{Effort}
& \textbf{Image Tool} & \textbf{Released} \\
\midrule
\multicolumn{5}{@{}l}{\textit{Open-weight}} \\
\midrule
DeepSeek-V4-Pro~\citep{deepseekai2026deepseekv4highlyefficientmilliontoken}
    & Hermes Agent & high & Nano Banana & 2026-08 \\
GLM-5.3~\citep{glm5team2026glm5vibecodingagentic}
    & Hermes Agent & high & Nano Banana & 2026-08 \\
Kimi-K3~\citep{kimiteam2026kimik3openfrontier}
    & Hermes Agent & high & Nano Banana & 2026-07 \\
\midrule
\multicolumn{5}{@{}l}{\textit{Closed-source}} \\
\midrule
Gemini 3.5 Flash~\citep{gemini35flash2026}
    & Gemini-CLI & high  & Nano Banana & 2026-05 \\
Claude Opus 4.8~\citep{frontier-claude-opus-4-8}
    & Claude Code & xhigh & Nano Banana & 2026-05 \\
GPT-5.6 Luna~\citep{OpenAI2026GPT56}
    & Codex-CLI  & xhigh & GPT-Image  & 2026-07 \\
GPT-5.6 Sol~\citep{OpenAI2026GPT56}
    & Codex-CLI  & xhigh & GPT-Image  & 2026-07 \\
\bottomrule
\end{tabular}
\end{table}

\begin{table}[h]
\centering
\small
\caption{
The two game docker environments used at \framework.}
\label{tab:docker}
\begin{tabular}{l l l}
\toprule
& \textbf{Minecraft} & \textbf{Terraria} \\
\midrule
Game / server  & Minecraft 1.21.1                 & tModLoader v2026.05.3.0 \\
Mod loader     & Fabric loader 0.19.3             & tModLoader v2026.05.3.0 \\
Build tool     & Gradle 8.12                      & tML-build \\
Runtime        & Java 21                          & bundled .NET 8.0 \\
Mod API        & Fabric API 0.116.14              & tModLoader API \\
Probe channel  & RCON+mineflayer                             & tModLoader ModSystems \\
\bottomrule
\end{tabular}%
\end{table}

\subsection{\bench{} Setup}
\label{app:igmbench_setup}

\begin{figure*}[t]
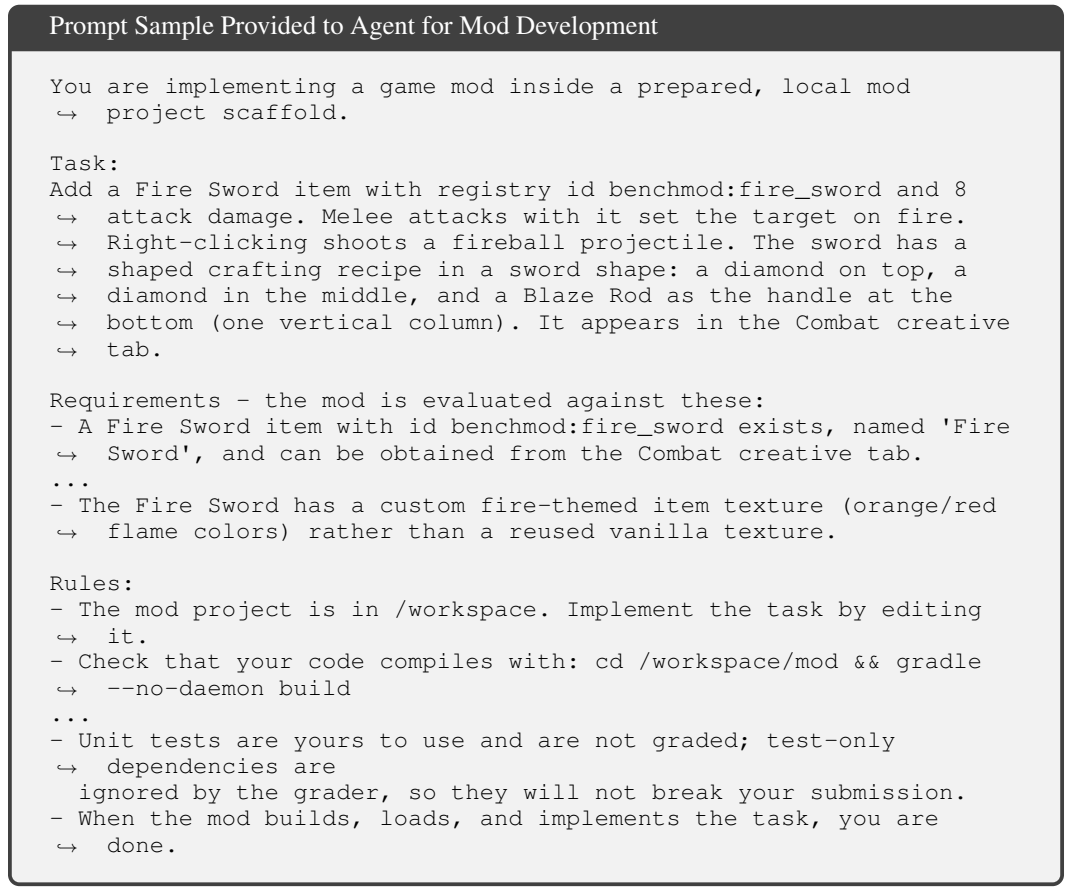

\centering
\begin{tcolorbox}[title={Prompt Sample Provided to Agent for Mod Development}]
\begin{Verbatim}[breaklines=true, breakanywhere=true,
  fontsize=\small]
You are implementing a game mod inside a prepared, local mod project scaffold.

Task:
Add a Fire Sword item with registry id benchmod:fire_sword and 8 attack damage. Melee attacks with it set the target on fire. Right-clicking shoots a fireball projectile. The sword has a shaped crafting recipe in a sword shape: a diamond on top, a diamond in the middle, and a Blaze Rod as the handle at the bottom (one vertical column). It appears in the Combat creative tab.

Requirements - the mod is evaluated against these:
- A Fire Sword item with id benchmod:fire_sword exists, named 'Fire Sword', and can be obtained from the Combat creative tab.
...
- The Fire Sword has a custom fire-themed item texture (orange/red flame colors) rather than a reused vanilla texture.

Rules:
- The mod project is in /workspace. Implement the task by editing it.
- Check that your code compiles with: cd /workspace/mod && gradle --no-daemon build
...
- Unit tests are yours to use and are not graded; test-only dependencies are
  ignored by the grader, so they will not break your submission.
- When the mod builds, loads, and implements the task, you are done.
\end{Verbatim}
\end{tcolorbox}
\caption{
Example prompt provided to the agent for an example world-editing task.
The prompt specifies the requested modification, evaluation-facing behavioral
requirements, project location, build command, and generic execution rules.
}
\label{fig:agent_prompt_example}
\end{figure*}

\begin{figure*}[t]
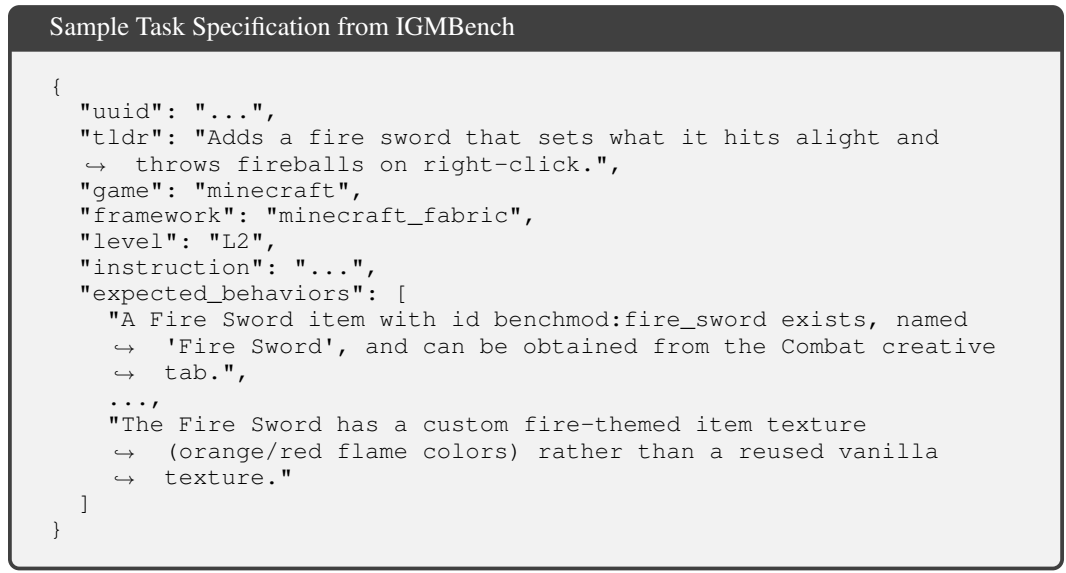

\centering
\begin{tcolorbox}[title={Sample Task Specification from \bench{}}]
\begin{Verbatim}[breaklines=true, breakanywhere=true, fontsize=\small]
{
  "uuid": "...",
  "tldr": "Adds a fire sword that sets what it hits alight and throws fireballs on right-click.",
  "game": "minecraft",
  "framework": "minecraft_fabric",
  "level": "L2",
  "instruction": "...",
  "expected_behaviors": [
    "A Fire Sword item with id benchmod:fire_sword exists, named 'Fire Sword', and can be obtained from the Combat creative tab.",
    ...,
    "The Fire Sword has a custom fire-themed item texture (orange/red flame colors) rather than a reused vanilla texture."
  ]
}
\end{Verbatim}
\end{tcolorbox}
\caption{
Sample task specification from \bench{}.
The task specification records benchmark metadata, the intervention level,
the natural-language request, and the evaluation criteria describing the
desired world edit.
}
\label{fig:sample_task_spec}
\end{figure*}

We show the agent prompt and task specification, also the evaluation sample from \bench{} in Figure~\ref{fig:agent_prompt_example}, Figure~\ref{fig:sample_task_spec}, and Figure~\ref{fig:sample_eval_spec}.

\subsection{Evaluating World-Edit Correctness}
\label{app:worldedit_eval}

Each behavior is described by seven fields: actor, action, subject, provenance, context, timing, and outcome. These specify who performs or causes the behavior, what action occurs, what it affects, its required causal history, the surrounding conditions, when it occurs, and its observable result. For example, ``a player harvesting a mature crop in Spring has a 20\% chance of receiving a Spring token" identifies the player as the actor, harvesting as the action, the mature crop as the subject, and Spring as the context. The outcome is a 20\% token drop chance per eligible harvest. Provenance requires evidence that the token came from the harvest rather than being supplied by the test. No additional timing constraint is specified. Each test has five components: fixture, trigger, measurement, acceptance rule, and prerequisite gates. The fixture establishes the actor, subject, and context. The trigger performs the required action. The measurement records the observable result, and the acceptance rule defines what counts as a pass. Prerequisite gates establish the conditions needed for a valid measurement. Timing requirements govern execution and observation, while provenance requirements govern setup and evidence. Compound requirements are split into separate checks for claims such as item registration, crafting, activation, cooldown, and appearance. For behaviors that modify a measurable property of the game world, tests compare that property before and after the required action. Other requirements are tested by inspecting properties directly or examining event traces. The components above define what each check must establish, while their
implementation depends on the game’s available interfaces. As shown in
Table~\ref{tab:reeval-timings}, the evaluation remains tractable at benchmark
scale, with an overall combined p95 of 508.1\,s per task and approximately
6 hours for a full benchmark run with five parallel workers. We next describe
how Minecraft and Terraria each perform test execution and measurement.

\begin{table}[h]
\centering
\small
\caption{
Evaluation runtime per task, reported as the 95th percentile in seconds.
Measurements were collected on a machine with two AMD EPYC 7302 processors
(32 cores total) and 503\,GiB of RAM, excluding build and orchestration
overhead. Using the overall combined p95 of 508.1\,s gives a conservative
sequential budget of 15.52 hours for all 110 tasks. In practice, evaluating
the full benchmark with five parallel workers takes approximately 6 hours.
}
\label{tab:reeval-timings}

\begin{tabular}{llrrrrr}
\toprule
Game & Phase & L1 & L2 & L3 & L4 & All levels \\
\midrule
 & Fresh-world load & 46.2 & 44.4 & 45.2 & 44.2 & 44.8 \\
Minecraft & Behavior checks & 236.1 & 582.8 & 421.1 & 1,738.1 & 632.9 \\
 & Combined & 269.6 & 615.9 & 463.8 & 1,770.2 & 670.2 \\
\midrule
 & Fresh-world load & 35.3 & 35.1 & 34.5 & 35.7 & 35.2 \\
Terraria & Behavior checks & 80.9 & 79.0 & 68.0 & 68.0 & 78.3 \\
 & Combined & 116.0 & 110.5 & 102.5 & 101.4 & 112.1 \\
\midrule                                            & Fresh-world load & 44.4 & 43.6 & 43.8 & 42.9 & 43.7 \\  
Overall & Behavior checks & 125.0 & 165.9 & 378.4 & 663.2 & 475.7 \\                                 & Combined & 164.6 & 207.8 & 418.2 & 698.4 & 508.1 \\     \bottomrule                                                   \end{tabular}

\end{table}

\subsubsection{Minecraft Deterministic Checks Design}

\textbf{Execution and Observation Interface.} Minecraft tests use Remote Console (RCON), Mineflayer~\citep{prismarinejs_mineflayer}, and a Fabric client~\citep{fabric2018}.
  RCON sets up fixtures and queries server state. Mineflayer
  performs ordinary player actions. The Fabric client handles
  actions that depend on client entrypoints, custom input
  handling, the candidate mod's client code, or graphical
  interfaces. Evaluator-owned helpers support test setup and
  observation.

\textbf{Runtime Subject and Interface Discovery.} Mineflayer discovers registry identifiers through command
  completion, removes duplicates, and excludes baseline
  identifiers. It converts identifiers and search terms into
  tokens, then selects candidates containing all required
  tokens and no excluded tokens. The Fabric client reads its
  loaded item registry directly. It matches substrings after
  lowercasing and removing non-alphanumeric characters, with
  optional namespace restrictions and filters. Both methods
  require exactly one match.When command syntax is unspecified, the evaluator
  reconstructs paths from the command tree and matches them
  against task-specific regular expressions. Named argument
  placeholders are filled with test-supplied values and
  checked against supported argument constraints. Resolution
  requires exactly one executable invocation, without
  expanding redirects. Discovery identifies how to invoke an
  interface; subsequent measurements establish whether it
  produces the required behavior.

\textbf{Gameplay Actions and Measurements} RCON sets up the actor, subject, environment, and required resources. Mineflayer or the Fabric client then performs the player action. RCON queries or client observations measure changes, final states, or events, which the evaluator checks against the acceptance rule. For requirements relative to vanilla gameplay, the evaluator first measures the same behavior without the candidate mod under matched test conditions.

\textbf{Timing and Transient Effects.} Timing requirements specify when to act and observe. Where supported, the evaluator advances or monitors server ticks to control delays and observation windows. Client-dependent actions also require confirmation that the interaction occurred. Short-lived status effects may expire or lose duration before a query. For supported checks, an evaluator-owned server helper records the maximum remaining duration observed within a monitoring interval. The fixture starts without the target effect to distinguish a new application from an existing effect. Cooldown checks attempt the action before and after the expected cooldown expires and check the outcome. Other timing checks use timed observations or event traces to assess duration and ordering.
  
\textbf{Custom Input and GUI Interaction.}
  The Fabric client performs actions involving client-side mod code, custom key bindings, or graphical interfaces. Each test specifies the interaction procedure. For clickable widgets, the evaluator matches active, visible controls by label using a task-specific regular expression, then calls the screen's existing input handling method at the selected widget's center. For custom-drawn interfaces, evaluator inject evaluator-owned hooks into supported rendering methods to record text, item identities, and coordinates. The evaluator clicks matching text or a nearby item icon. Text-item associations rely on geometric constraints. The evaluator inspects container slots through the current screen renderer's method, distinguishing them from player-inventory slots and reading their contents and insertion rules. To insert an item, it uses slot-click operations to move a test-specified hotbar stack into the first empty container slot that accepts it. To collect output, it transfers a matching stack from a slot that disallows insertion to an empty, test-specified hotbar slot. For key-driven interactions, the evaluator matches task-specific terms against registered key-binding names, labels,
  and categories. It then generates a press notification for
  the assigned key, allowing the candidate mod to process the
  input through its registered handler.

\textbf{Probabilistic Mechanics.} 
There are four main ways in testing requirements entailing stochastic requirement like "a player harvesting a mature crop in Spring has a 20\% chance to receive the Spring token"
\begin{enumerate}
    \item Monte Carlo-style sampling. The evaluator repeats an eligible event for a task-dependent number of trials and compares outcome counts against predefined acceptance bounds. Qualitative requirements instead use occurrence checks or comparisons.

    \item Direct parameter inspection. The evaluator reads registered parameters, such as spawn weights, from runtime server state.

    \item Two-point boundary testing. For supported threshold-based decisions, the evaluator forces random inputs immediately below and at the required threshold, then inspects the outcomes.

    \item Bounded enumeration. The evaluator executes all supported random-choice paths within its enumeration limits and weights their observed outcomes to calculate an expectation.
\end{enumerate}

\subsubsection{Terraria Deterministic Checks Design}

  \textbf{Execution and Observation Interface.}
  Terraria tests use an evaluator mod from tModLoader~\citep{tmodloader} to setup fixtures, triggers and measurer. The measurements are then emitted to an external verification engine to compare against the acceptance rules. 

  \textbf{Runtime Subject Discovery.}
 The evaluator mod reads loaded content registries directly and locates the relevant subjects using a name-matching process similar to Minecraft's.

  \textbf{Probabilistic Mechanics.}
 Supported stochastic checks use fixed-budget repeated
  trials, following the Monte Carlo-style sampling described for Minecraft. Terraria's crafting checks also examine joint outcomes within each trial. For example, if ingredients A and B each independently have a 50\% chance of being saved, the four combinations of saving and consuming them should each occur with probability 25\%. 

\subsection{Evaluating Visual Correctness}
\label{app:visual_eval}

Our visual evaluation measures whether agent-produced assets both depict the
requested content and remain compatible with the visual distribution of the
host game. As described in the main paper, we use category-conditioned TPIPS
with separate text factors for semantic identity and art style. For each asset
category, the pass threshold is calibrated from leave-one-out scores over
original game assets, accounting for differences in visual variability across
categories. We use $\alpha=0.15$ and $k=5$ throughout the main experiments.
Here, we validate these choices through a preliminary human study and
sensitivity analysis.

\begin{figure}[h]
\centering
\includegraphics[width=0.95\textwidth]{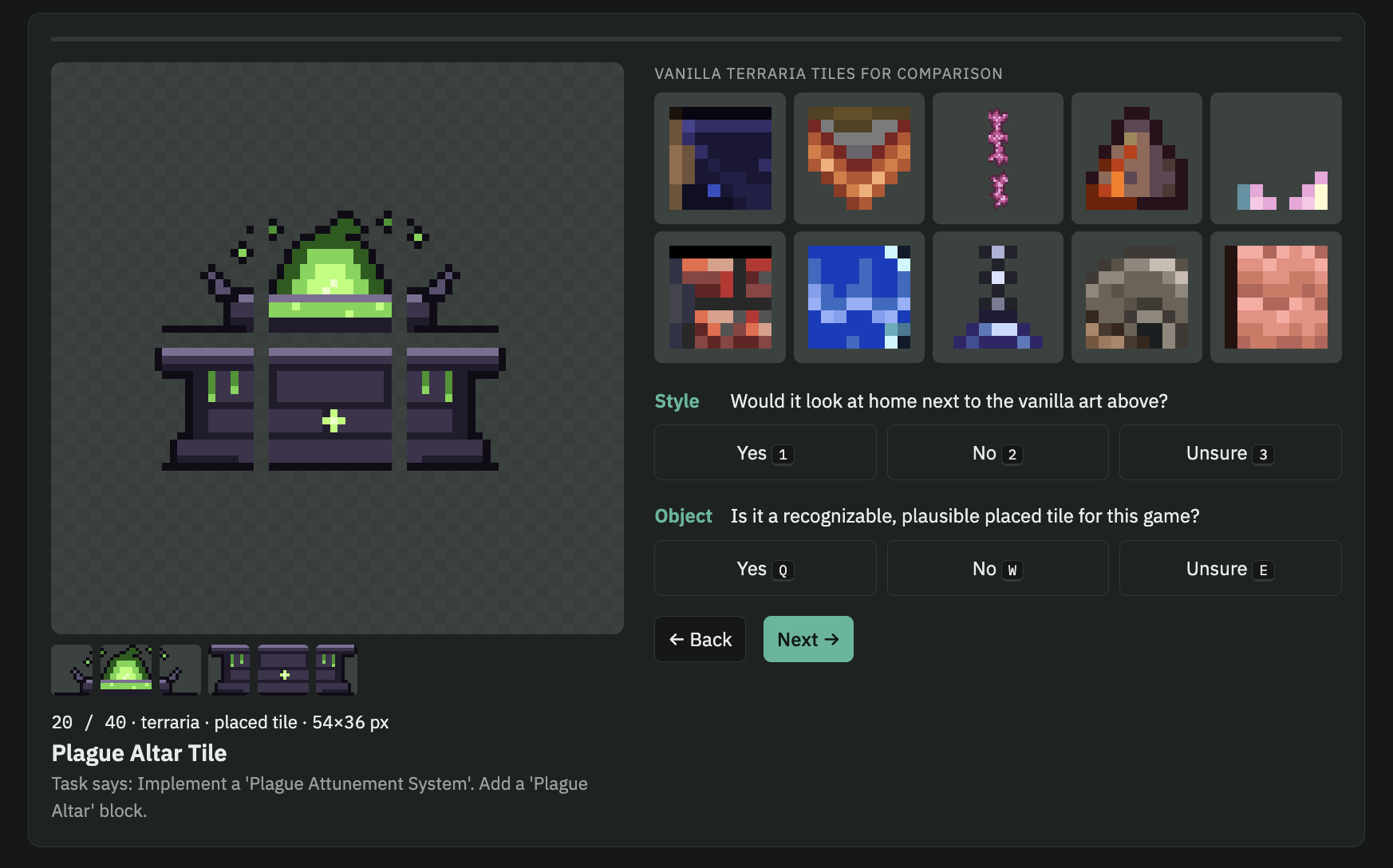}
\caption{
Interface used for the preliminary human perceptual study.
Raters judged whether agent-produced assets were stylistically compatible
with the shown vanilla references and semantically plausible for the
requested object category.
}
\label{fig:app_human_texture_study}
\end{figure}

\begin{table}[h]
\centering
\small
\caption{
Calibration of the pass threshold $\alpha$ using the preliminary human study.
Nine raters judged 40 agent-produced textures blind to metric scores.
\emph{Acc.} is the fraction of majority-accepted textures passed by the
metric, \emph{Rej.} the fraction of majority-rejected textures rejected by
the metric, and \emph{Bal.} their mean. \emph{Vanilla} is the fraction of
original game assets passing both checks. The shaded row is the setting used
in the main experiments.
}
\label{tab:app_visual_alpha}

\begin{tabular}{c ccc ccc c c}
\toprule
 & \multicolumn{3}{c}{\textbf{Art Style} vs.\ style votes} & \multicolumn{3}{c}{\textbf{Semantic} vs.\ object votes} & & \\
\cmidrule(lr){2-4}\cmidrule(lr){5-7}
$\alpha$ & Acc. & Rej. & Bal. & Acc. & Rej. & Bal. & \textbf{Mean Bal.} $\uparrow$ & Vanilla \\
\midrule
0.01 & 0.92 & 0.09 & 0.51 & 0.90 & 0.22 & 0.56 & 0.53 & 0.98 \\
0.02 & 0.88 & 0.27 & 0.58 & 0.80 & 0.33 & 0.57 & 0.57 & 0.97 \\
0.05 & 0.73 & 0.45 & 0.59 & 0.63 & 0.33 & 0.48 & 0.54 & 0.92 \\
0.10 & 0.58 & 0.64 & 0.61 & 0.53 & 0.56 & 0.54 & 0.58 & 0.85 \\
\rowcolor{gray!15} 0.15 & 0.38 & 0.82 & 0.60 & 0.47 & 0.78 & 0.62 & 0.61 & 0.78 \\
0.20 & 0.27 & 0.91 & 0.59 & 0.40 & 0.78 & 0.59 & 0.59 & 0.72 \\
0.25 & 0.15 & 0.91 & 0.53 & 0.30 & 1.00 & 0.65 & 0.59 & 0.67 \\
\bottomrule
\end{tabular}
\end{table}

\textbf{Human calibration of the pass threshold.}
We conducted a preliminary human perceptual study to guide the choice of
$\alpha$. Figure~\ref{fig:app_human_texture_study} shows the annotation
interface. Nine raters evaluated 40 agent-produced textures without access to
the automatic metric scores. For each texture, raters judged whether the asset
would visually fit alongside the shown vanilla game art and whether it
represented a recognizable and plausible object of the requested type.
Table~\ref{tab:app_visual_alpha} compares the majority human judgments with
the automatic style and semantic verdicts across candidate values of
$\alpha$. Because the human labels are skewed toward acceptance, we use
balanced accuracy rather than raw agreement. The setting $\alpha=0.15$
achieves the highest mean balanced accuracy across the two checks. We use this
setting throughout the main evaluation. This study is intended as a
calibration sanity check rather than a definitive perceptual benchmark, as
the sample is small and inter-rater agreement is modest (Krippendorff's alpha~\citep{Krippendorff2011ComputingKA} equals 0.25 for style and 0.20 for object.)

\textbf{Sensitivity to the number of neighbours.}
We additionally evaluate sensitivity to the neighbourhood size $k$.
For each value of $k$, the category-specific threshold is recalibrated using
the same leave-one-out procedure, so differences reflect only the choice of
neighbourhood size. As shown in Table~\ref{tab:app_visual_k}, the resulting
texture rankings and pass/fail decisions remain stable over a broad range of
$k$. Relative to $k=5$, Spearman rank correlation remains at least $0.94$
for all tested values, while at least 88\% of evaluated textures retain the
same verdict across the full range. Model-level ordering is also largely
preserved. We therefore use $k=5$ as a middle setting that avoids relying on
a single nearest neighbour while limiting the influence of less relevant
references in smaller categories.

\begin{table}[h]
\centering
\small
\caption{
Sensitivity to the neighbour count $k$ at $\alpha=0.15$ over 1,760 scored
agent textures in sprite-like categories. Thresholds are recalibrated by
leave-one-out for each $k$. \emph{Pass} is the agent pass rate,
\emph{Rank corr.} is the Spearman correlation of texture rankings with
$k=5$, \emph{Same verdict} is the fraction retaining the same pass/fail
decision, and \emph{Model order} is the Spearman correlation of model-level
pass rates with those at $k=5$. The shaded rows indicate the setting used in
the main experiments.
}
\label{tab:app_visual_k}

\begin{tabular}{l c c c c c}
\toprule
\textbf{Metric} & $k$ & \textbf{Pass} & \textbf{Rank corr.} & \textbf{Same verdict} & \textbf{Model order} \\
\midrule
Art Style & 1 & 0.32 & 0.948 & 0.91 & 0.95 \\
 & 3 & 0.32 & 0.993 & 0.97 & 1.00 \\
\rowcolor{gray!15}  & 5 & 0.33 & 1.000 & 1.00 & 1.00 \\
 & 10 & 0.37 & 0.989 & 0.94 & 0.98 \\
 & 20 & 0.42 & 0.969 & 0.88 & 1.00 \\
\addlinespace
Semantic & 1 & 0.37 & 0.943 & 0.88 & 0.93 \\
 & 3 & 0.39 & 0.993 & 0.95 & 0.90 \\
\rowcolor{gray!15}  & 5 & 0.41 & 1.000 & 1.00 & 1.00 \\
 & 10 & 0.45 & 0.988 & 0.94 & 0.98 \\
 & 20 & 0.49 & 0.960 & 0.88 & 0.98 \\
\bottomrule
\end{tabular}
\end{table}

\clearpage

\begin{figure*}[t]
\centering
\begin{tcolorbox}[title={Sample Evaluation Specification from \bench{}}]
\begin{Verbatim}[breaklines=true, breakanywhere=true, fontsize=\small]
{
  "verify": {
    "setup": [
      {
        "rcon": "kill @e[type=minecraft:item]"
      },
      {
        "rcon": "setblock 0 -59 0 minecraft:chest"
      },
      ...,
      {
        "rcon": "time set midnight"
      },
      {
        "bot": {
          "op": "physics",
          "enabled": false
        }
      },
      ...,
      {
        "rcon": "scoreboard players operation #fireball_damage firesword -= #fireball_after firesword"
      }
    ],
    "checks": [
      ...,
      {
        "name": "the fired fireball damages a target it hits",
        "rcon": "scoreboard players get #fireball_damage firesword",
        "rule": {
          "min": 1
        },
        "kind": "positive",
        "depends_on": [
          "benchbot fixture is present after setup"
        ],
        "tags": [
          "BOT_RCON_NOW"
        ]
      }
    ]
  }
}
\end{Verbatim}
\end{tcolorbox}
\caption{
Sample evaluation specification from \bench{}.
The verification view defines the controlled game setup and executable checks
used to determine whether the requested world edit is correctly realized.
}
\label{fig:sample_eval_spec}
\end{figure*}

\clearpage
\section{\bench{} Diversity}
\label{app:diversity}

\begin{figure}[h]
\centering
\includegraphics[width=0.8\textwidth]{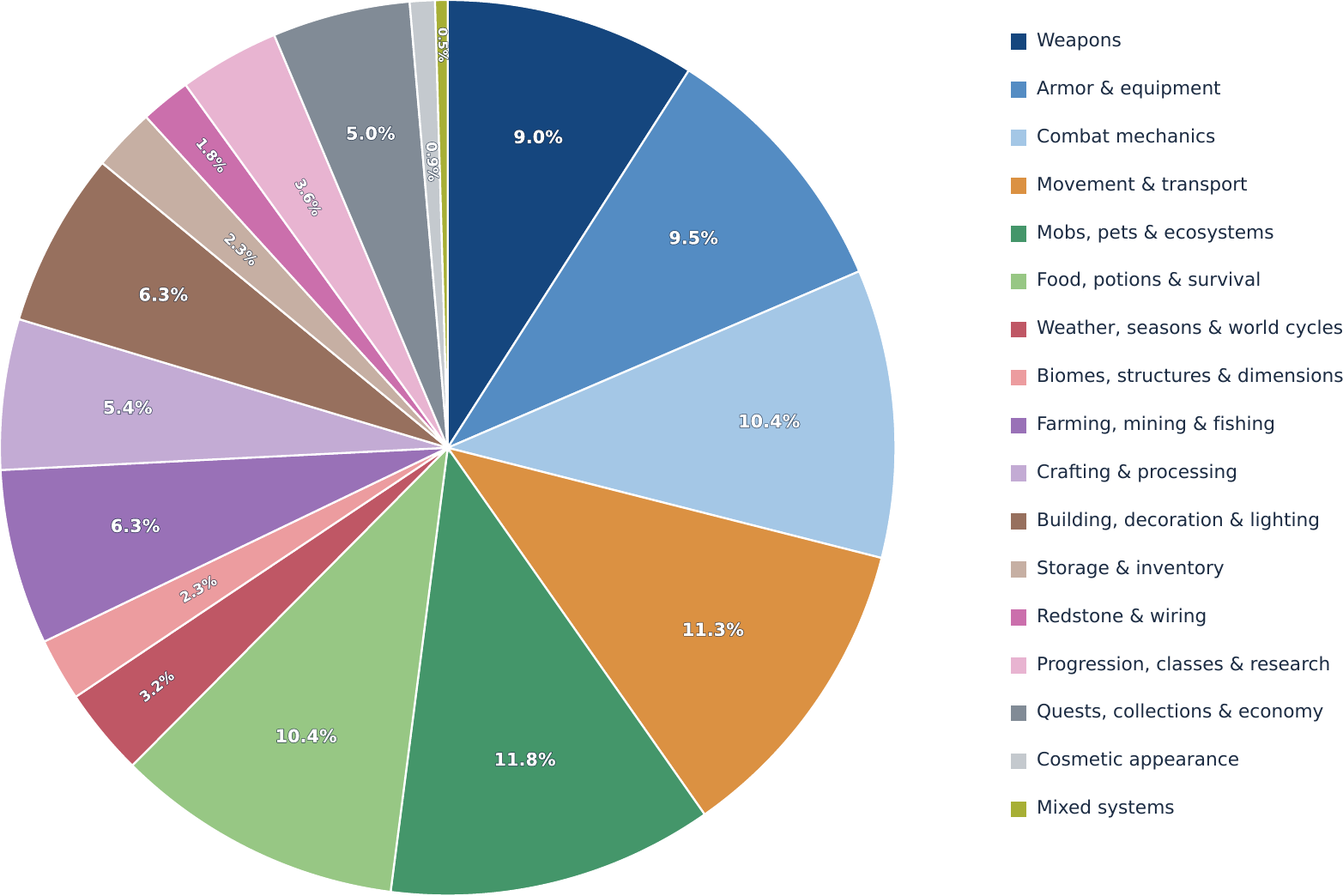}
\caption{Task genre distribution across all 110 tasks in \bench{}.
Percentages indicate the share of tasks in each genre.}
\label{fig:app_task_genre_total}
\end{figure}

\begin{figure}[h]
\centering
\begin{minipage}[t]{0.48\textwidth}
  \centering
  \includegraphics[width=\linewidth]{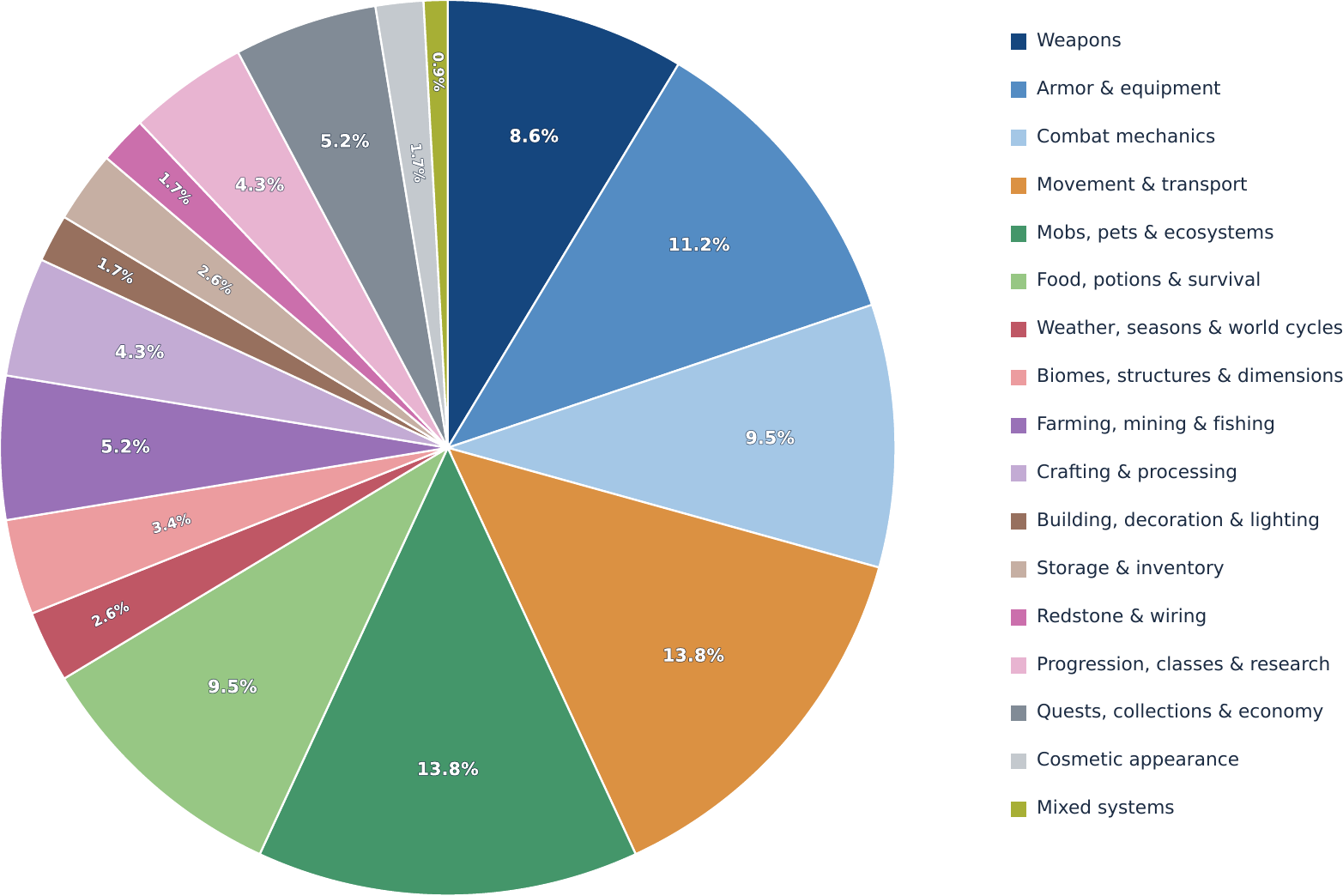}
  \par\smallskip
  (a) Minecraft
\end{minipage}\hfill
\begin{minipage}[t]{0.48\textwidth}
  \centering
  \includegraphics[width=\linewidth]{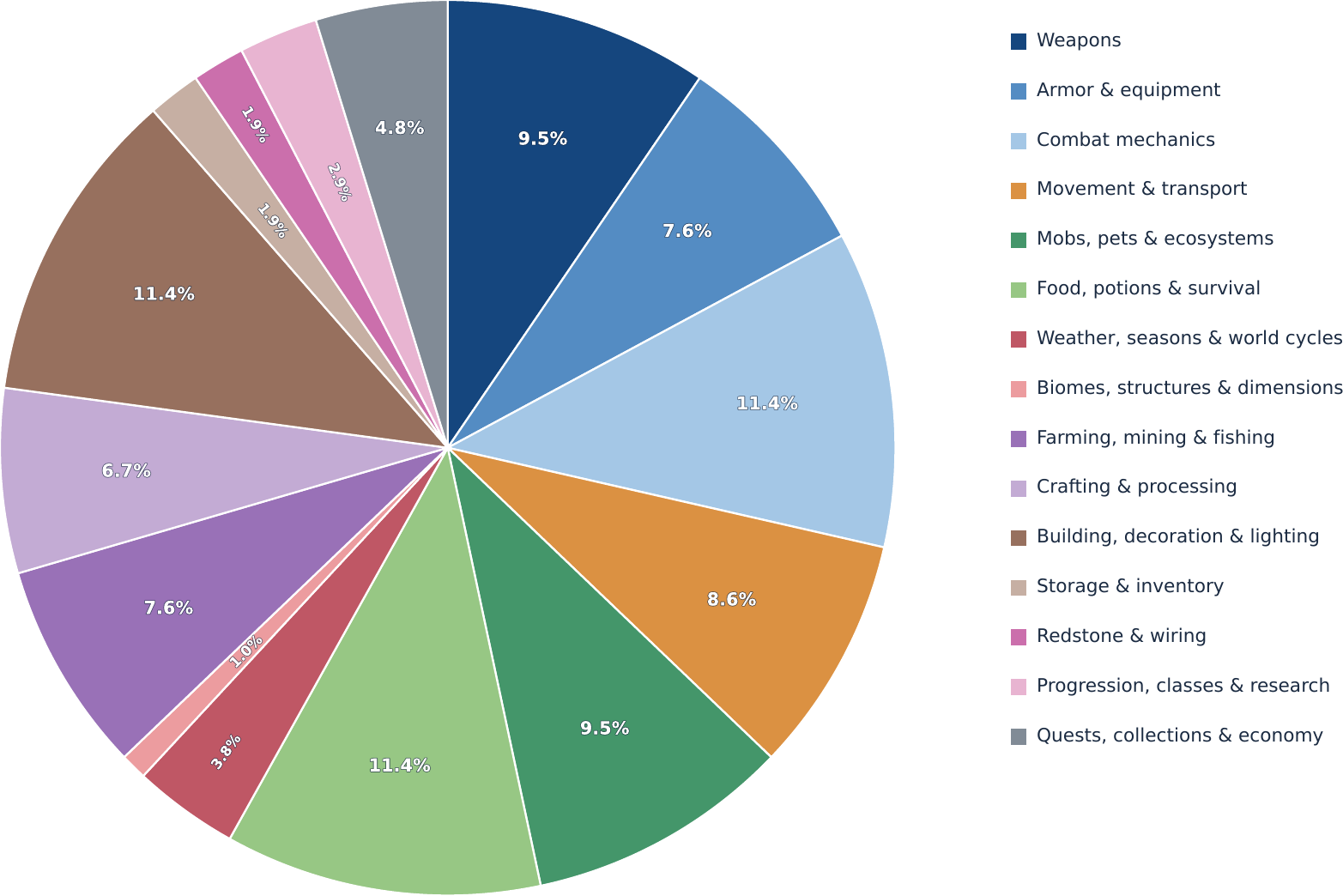}
  \par\smallskip
  (b) Terraria
\end{minipage}
\caption{Task genre distributions in \bench{} by game.
(a) Minecraft (57 tasks); (b) Terraria (53 tasks).
Percentages are computed within each game.}
\label{fig:app_task_genres_by_game}
\end{figure}

Figure~\ref{fig:app_task_genre_total} summarizes the overall task-genre
distribution in \bench{}, while Figure~\ref{fig:app_task_genres_by_game}
shows the corresponding distributions for Minecraft and Terraria.

\end{document}